\documentclass[journal]{IEEEtran}
\usepackage{amsmath,amsfonts}
\usepackage{algorithmic}
\usepackage{algorithm}
\usepackage{array}
\usepackage[caption=false,font=normalsize,labelfont=sf,textfont=sf]{subfig}
\usepackage{textcomp}
\usepackage{stfloats}
\usepackage{url}
\usepackage{verbatim}
\usepackage{graphicx}
\usepackage{paralist}
\usepackage{color}
\usepackage{multirow}
\usepackage{booktabs}
\usepackage[hidelinks]{hyperref}

\def \ie {\emph{i.e.}~}
\def \eg {\emph{e.g.}~}
\def \etc {\emph{etc.}~}
\def \etal {\emph{et al.}~}

\def \wrt {\emph{w.r.t.}~}

\newcommand{\miaojing}[1]{{#1}}

\newcommand{\shi}[1]{{#1}}

\begin{document}

\title{Redesigning Multi-Scale Neural Network for Crowd Counting}

\author{Zhipeng Du,~%~\IEEEmembership{Staff,~IEEE,}
        Miaojing Shi$^*$,~\IEEEmembership{Senior Member,~IEEE,}
        Jiankang Deng,~\IEEEmembership{Member,~IEEE,}
        Stefanos Zafeiriou%~\IEEEmembership{Member,~IEEE,}
        % <-this % stops a space
% \thanks{Z. Du and M. Shi are with the Department of Informatics, King's College London. E-mails: \tt\small \{zhipeng.du, miaojing.shi\}@kcl.ac.uk.}% <-this % stops a space
\thanks{$^*$Corresponding author}
\thanks{Z. Du is with the Department of Informatics, King's College London. E-mail: \tt\small zhipeng.du@kcl.ac.uk.}
\thanks{{M. Shi is with the College of Electronic and Information Engineering, Tongji University. E-mail: \tt\small mshi@tongji.edu.cn.}}% <-this % stops a space
\thanks{J. Deng and S. Zafeiriou are with the Department of Computing, Imperial College London. E-mails: \tt\small\{j.deng16, s.zafeiriou\}@imperial.ac.uk.}}

% The paper headers
% \markboth{IEEE Transactions on Image Processing}{}

%\IEEEpubid{0000--0000/00\$00.00~\copyright~2021 IEEE}
% Remember, if you use this you must call \IEEEpubidadjcol in the second
% column for its text to clear the IEEEpubid mark.

\maketitle

\begin{abstract}
Perspective distortions and crowd variations make crowd counting a challenging task in computer vision. To tackle it, many previous works have used multi-scale architecture in deep neural networks (DNNs). Multi-scale branches can be either directly merged (\eg by concatenation) or merged through the guidance of proxies (\eg attentions) in the DNNs. Despite their prevalence, these combination methods are not sophisticated enough to deal with the per-pixel performance discrepancy over multi-scale density maps. In this work, we redesign the multi-scale neural network by introducing a hierarchical mixture of density experts, which hierarchically merges multi-scale density maps for crowd counting. Within the hierarchical structure, an expert competition and collaboration scheme is presented to encourage contributions from all scales; pixel-wise soft gating nets are introduced to provide pixel-wise soft weights for scale combinations in different hierarchies. The network is optimized using both the crowd density map and the local counting map, where the latter is obtained by local integration on the former. Optimizing both can be problematic because of their potential conflicts. We introduce a new relative local counting loss based on relative count differences among hard-predicted local regions in an image, which proves to be complementary to the conventional absolute error loss on the density map. Experiments show that our method achieves the state-of-the-art performance on five public datasets, \ie ShanghaiTech, UCF\_CC\_50, JHU-CROWD++, NWPU-Crowd and Trancos. Our codes will be available at \emph{\color{magenta}{https://github.com/ZPDu/Redesigning-Multi-Scale-Neural-Network-for-Crowd-Counting}}.

\end{abstract}


\section{Introduction}
\label{sec:intro}
Crowd counting in the computer vision field automatically counts people in images. The rapid growth of the world’s population has resulted in more frequent crowd gatherings, making crowd counting a fundamental task for crowd control.

Current state of the arts in crowd counting employ a deep neural network (DNN) to estimate a density map of an image where the integral over the density map gives the total count~\cite{zhang2016cvpr}. The main challenge lies in the drastic change of crowd density and perspective both within and across images. One commonly used technique to address this issue is the multi-scale architecture, where multiple density maps are estimated from different scales of the DNN. The scales can be implemented in the form of multi-column~\cite{zhang2016cvpr,onoro2016eccv,sam2017cvpr,ranjan2018eccv} or multi-branch~\cite{boominathan2016mm,zhang2018wacv,liu2019iccv}. To combine them, many previous works have concatenated or added multi-scale outputs in the network~\cite{zhang2016cvpr,boominathan2016mm,onoro2016eccv,zhang2018wacv}. Recent works have improved this by leveraging proxies such as image perspective~\cite{shi2019cvpr,yan2019iccv,yang2020cvpr} and  attention~\cite{shi2019iccv,jiang2020cvpr} as the local guidance. The performance discrepancy over multi-scale density maps is implicitly assessed via these proxies, which primarily vary over regions (\eg\ crowd clusters, perspective bands).

\begin{figure}[t]
\begin{center}
% \begin{overpic} 
% [width=\linewidth]
% {example-image-a}
% \end{overpic}
\includegraphics[width=\linewidth]{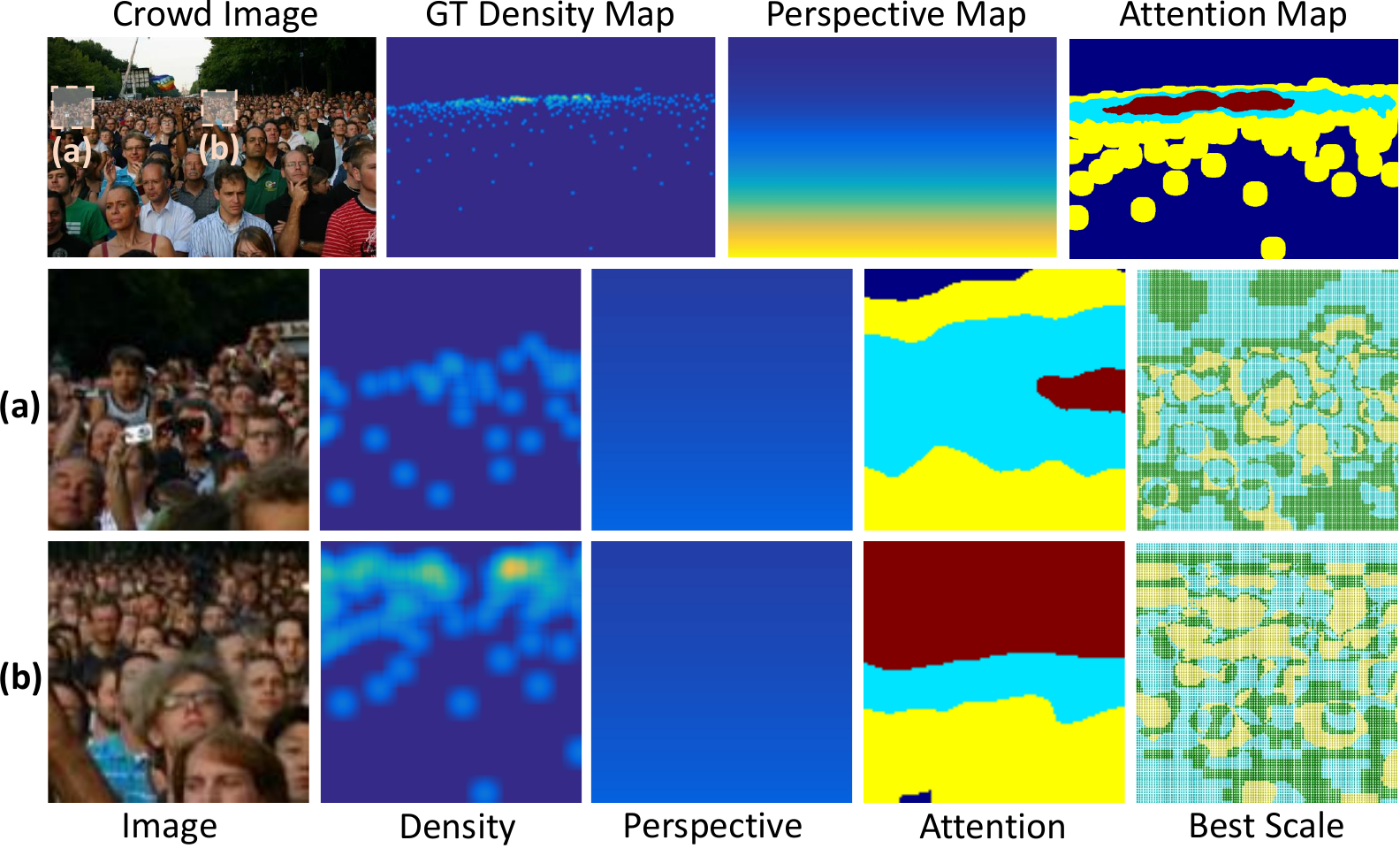}
\end{center}

\caption{
\miaojing{
First row: a crowd image and its corresponding ground truth density map, perspective map, and attention map. In the perspective map, colours from yellow to navy blue denote perspective values from highest to lowest; in the attention map, colours from brown, cyan, yellow and navy blue denote attention values from highest to lowest. Second row (rightmost): the best-performing scale map obtained by running a three-scale crowd counting model (see Sec.~\ref{sec:implementation}: Baseline) on the image; we use faint blue, yellow, and green to represent the three scales, respectively.  
% The left column is an image with its ground truth density map where we sample two local regions and place them in the middle column. We run a three-scale crowd density estimation model (see Sec.~\ref{sec:implementation}: Baseline) on this image and illustrate the best performing scale in the two local regions in the right column. Blue, yellow, and green represent each scale, respectively.
} }
%It's obvious that one scale does not always perform well on a similar distribution.}
\label{fig:teaser}

\end{figure}

To illustrate the per-pixel performance discrepancy over different scales, we train a three-scale crowd density estimation network based on an encoder-decoder architecture \shi{(see Sec. IV-B-Baseline)} and test it on an image in Fig.~\ref{fig:teaser}. 
%compare the ground truth density map to the estimated density map of each scale. At each pixel, we find the best performing scale that has the minimal difference (error) to the ground truth. We use yellow, pink and blue to represent the three scales, respectively and visualize the per-pixel best performing scale in the two selected patches in the bottom right of this figure. 
\miaojing{The estimated density map of each scale is compared to the ground truth density map. 
At each pixel, we find the best performing scale that has the minimal prediction error to the ground truth and visualize the per-pixel best performing scale. Meanwhile, we also follow \cite{shi2019cvpr,jiang2020cvpr} to generate the perspective map and attention map of this image, respectively. 
%per-pixel density value from the ground truth to predictions of three scales in the image. 
The figure shows that the distribution map of the best performing scale clearly differs from the commonly used perspective/attention map. The latter is normally utilized as a weighting scheme or selection criterion for multi-scale combination \cite{shi2019cvpr,jiang2020cvpr}, which apparently is inaccurate.} A finer feature steering scheme on a per-pixel basis is needed for better multi-scale combination. To address this issue, we use the \emph{mixture of experts} (MoE)~\cite{jacobs1991nn}.     

MoE is a classic ensemble learning method that divides the problem space among a number of experts supervised by a gating network. Since its introduction in 1990's~\cite{jacobs1991nn,jordan1994nn}, it has been implemented with multiple forms of experts, \eg support vector machine (SVM)~\cite{collobert2002nn}, Gaussian process~\cite{tresp2001nips}, and DNN~\cite{eigen2013arxiv}. It has also been used in a multi-scale DNN for crowd counting~\cite{kumagai2018MVA}, but in an old-fashioned pipeline supervised by global crowd counts. In light of recent advances in crowd counting, this paper redesigns the traditional MoE and introduces a new DNN-based two-level hierarchical mixture of density experts (HMoDE): in the first level, different experts (density outputs from different scales) are formed into groups and combined within each group; in the second level, all group outputs are combined together. Two main elements are highlighted: i) an
expert collaboration and competition scheme is introduced in the first level which allows group overlap and balances expert importance within each group to encourage the contribution from every expert; ii) the pixel-wise soft-gating nets are designed which produce per-pixel control on the combination of multi-scale density maps in the two levels. \shi{Compared to the perspective/attention maps in~\cite{shi2019cvpr,jiang2020cvpr}, our HMoDE is more suitable to address the per-pixel performance discrepancy in multi-scale combination.} 

%. Two gating nets are used for each of the two levels, respectively. In order to distinguish between their objectives, an attention module is further implemented on the second net.   

The multi-scale density outputs are optimized on per-pixel based density errors \wrt the ground truth. In addition, we also pay attention to the local count errors using local count estimation, an estimation target which was widely employed in traditional crowd counting methods~\cite{idrees2013cvpr,chen2012bmvc}. Its popularity has recently increased due to its robustness over noise, which is also believed to be more consistent with the evaluation criteria~\cite{liu2020eccv}. However, there is no substantial evidence showing that benefits can be obtained by optimizing on both crowd densities and local counts as they may be redundant or contradictory in different areas. Therefore, we aim to separate their optimization forces: in contrast to absolute errors utilized on the crowd density map, we focus on relative errors on the local counting map. Any two local regions in an image can form a ranking pair {by their local counts} yet only hard-predicted regions with large estimation errors are used to compare amongst each other. 
%We generate the local counting map from the density map by integrating density values within evenly partitioned regions. 
A novel relative local counting loss is introduced based on relative count differences among hard-predicted regions. The hard mining on regions is more robust than on pixels, giving theoretical support to our design. 

Overall, the contribution of this paper is three-fold:  
\begin{compactitem}
\item We introduce a \emph{hierarchical mixture of density experts} architecture which includes two key elements, the expert competition and collaboration scheme and the pixel-wise soft gating net. It offers a transformative way to redesign the density estimation network in crowd counting. 
\item We introduce a scheme of \emph{learning from relative local counting} which focuses on optimizing the relative local count errors among hard-predicted local regions. It is complementary to the conventional optimization on crowd density maps and is applicable to many other methods.   
\item Our method outperforms the state of the art with sizeable margins on standard benchmarks, ShanghaiTech~\cite{zhang2016cvpr}, UCF\_CC\_50~\cite{idrees2013cvpr}, JHU-CROWD++~\cite{sindagi2020jhu}, NWPU-Crowd~\cite{wang2020pami} and Trancos~\cite{guerrero2015ibpra}.
%demonstrate that our method improves the state of the art with a sizable margin.
\end{compactitem}
%Extensive experiments on standard benchmarks, ShanghaiTech~\cite{zhang2016cvpr}, UCF\_CC\_50~\cite{idrees2013cvpr}, JHU-CROWD++~\cite{sindagi2020jhu}, UCF-QNRF~\cite{idrees2018eccv} and Trancos~\cite{guerrero2015ibpra},  demonstrate that our method improves the state of the art with a sizable margin.  

\section{Related works}
\label{sec:related}
We survey crowd counting in three aspects: estimation targets, multi-scale architecture and mixture of experts. 

\subsection{Estimation targets in crowd counting}
%{Traditional methods} in crowd counting tend to directly predict crowd counts globally or locally in images~\cite{chan2009iccv,paul2017iccvw,lu2017plantmethod}.  
%the latter is more favoured~\cite{paul2017iccvw,lu2017plantmethod}. 
Traditional methods in crowd counting use handcrafted features such as edges and textures to predict crowd counts globally or locally in images~\cite{chan2009iccv,paul2017iccvw,lu2017plantmethod}. These features were later replaced by the deep representations in DNNs~\cite{chen2012bmvc,shang2016icip}. 
%For instance, Chen \etal\cite{chen2012bmvc} proposed a multi-output regressor to count pedestrians in different local regions. Shang \etal\cite{shang2016icip} utilized recurrent network layers to predict both local and global counts. 
In contrast to the crowd count estimation, Lempitsky \etal\cite{lempitsky2010nips} proposed to estimate a density map of a crowd image whose integral over the image gives the total count. A density map encodes the spatial information of pedestrians. Estimating it in a DNN is more robust than simply estimating the crowd count~\cite{zhang2016cvpr}.  It is thus the most commonly used estimation target in modern crowd counting~\cite{onoro2016eccv,sindagi2017iccv,li2018cvpr,xu2019iccv,zhao2022tmm,du2023aaai}. Recently, some approaches of crowd count estimation have become repopularised~\cite{xiong2019iccv,liu2020eccv}. For instance, Xiong \etal\cite{xiong2019iccv} utilized the theory of `divide-and-conquer' to construct a neural network which classifies local counts into different intervals. 
%Liu \etal\cite{liu2020eccv} proposed a new estimation target called local counting map and an adaptive mixture regression framework with better accuracy and more stable training phase.}
%Our work estimates a density map and generates a local counting map on top of it. The strengths of both are advantageously taken by applying a relative counting loss on the latter.
There
also exist some other estimation targets
in crowd counting. For instance, detection-based methods regress bounding boxes of pedestrians in crowd images~\cite{liu2018cvpr,lian2019cvpr,liushi2019cvpr, liu2020acmmm}. In addition, auxiliary information,~\eg multiple views~\cite{zhang2021cvpr,qiu2019icme}, depth images~\cite{lian2019cvpr},  thermal images~\cite{liu2021cvpr} and unlabelled images~\cite{liu2018bcvpr,cheng2019iccv,zhao2020eccv}, is also leveraged to help crowd counting.    %thermal information in and count value relations in unlabelled images~\cite{liu2018bcvpr,liu2019pami,cheng2019iccv} to help with predicting bounding boxes or density maps. Inspired by the thought that one region must contain fewer or equal individuals compared to its super-region, we introduce relative local counting loss to incorporate relative local counting information. 
Our work estimates a density map and generates a local counting map on top of it. For the latter, we introduce a relative local counting loss to compare the predicted counts between local regions. 
%which operates on a series of non-overlapping and hard-predicted regions and compare their predicted local counts.
~\miaojing{Some recent methods also exploit the relations between crowd counts from overlapped local regions in unlabeled images~\cite{liu2018bcvpr} or from global images without the supervision of person locations~\cite{yang2020eccv}. Different from them, our relative local counting operates on a series of non-overlapping and hard-predicted regions in labeled images.} Together with the conventional density estimation loss~\cite{zhang2016cvpr}, we 
%The strengths of both are advantageously taken by applying a relative counting loss on the latter.
for the first time enable a joint optimization on both the crowd density map and the local counting map. 
%Because of the availability of ground truth, it operates on a series of evenly squared non-overlap hard regions in the image, and compares the relative local counts among them. 
\begin{figure*}[t]
\begin{center}

\includegraphics[width=1\textwidth]{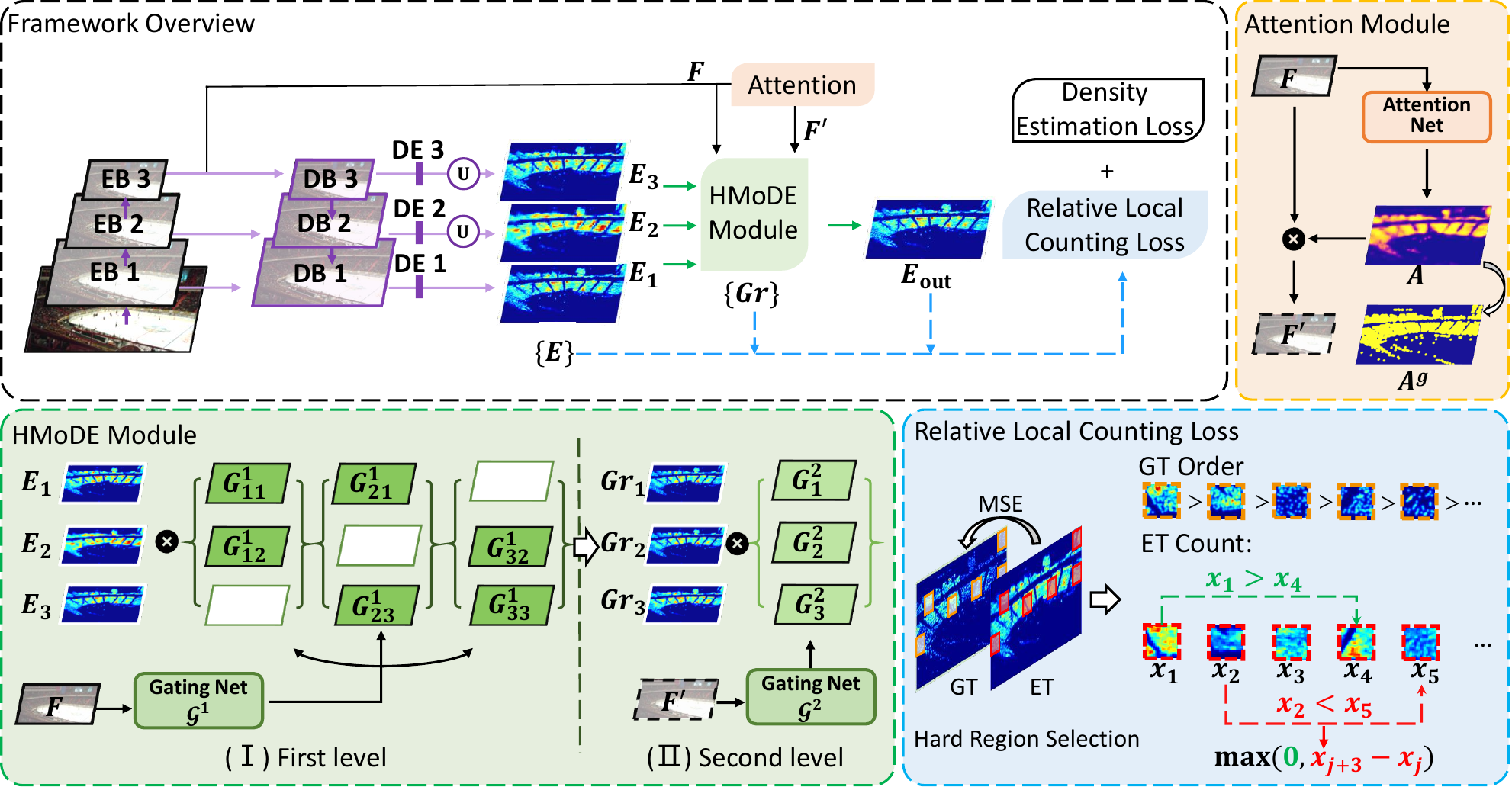}
\end{center}

\caption{
Overview of our framework with default setting (expert number $K = 3$, and group size $N = 2$). EB 1$\sim$3 are encoding stages and DB 1$\sim$3 are decoding stages. DE 1$\sim$3 are density estimation heads, and $E_1\sim E_3$ are density experts. The top left shows the workflow of our framework. The HMoDE module illustrates the controlling of the gating nets $\mathcal G^1/\mathcal G^2$ on the three density outputs in two levels: we group any two density experts (density maps) in the first level and all three density experts in the second level; density maps are pixel-wisely multiplied to corresponding weight maps in each group and then added together into one for this group. Blank weight maps indicate that the corresponding density maps are not present in certain groups. The top right shows the attention module, where the attention net takes the encoder feature $F$ as the input and generates the attention map $A$. $A$ is multiplied to $F$ to obtain the attention-guided feature $F'$. The attention net is supervised by the ground truth attention map $A^g$. The bottom right illustrates the proposed relative local counting loss, which operates on selected hard-predicted regions. This loss together with the conventional density estimation loss is applied to all intermediate and final density outputs.
}
\label{fig:overview}
\end{figure*}

\subsection{Multi-scale architecture}
As heavy occlusions and perspective distortions commonly occur in crowd images, multi-scale DNN architectures are often exploited in crowd counting, \miaojing{which can be implemented in the form of either multiple columns~\cite{zhang2016cvpr,onoro2016eccv,sam2017cvpr,ranjan2018eccv} or branches~\cite{boominathan2016mm,zhang2018wacv,cao2018eccv,liu2019iccv}. The multi-column structure uses multiple parallel sub-networks to extract multi-scale features from the input image where each sub-network separately processes the input at a certain scale. Multiple columns normally do not share the backbone. For instance, Zhang \etal\cite{zhang2016cvpr} {employed} a three-column network and concatenated the three-column features in the end for density map estimation; Onoro \etal\cite{onoro2016eccv} used a pyramid of image patches and fed them into different DNN columns. 
Sam~\etal~\cite{sam2017cvpr} processed each image patch using independent network column so as to cope with the density variations. In contrast, multi-branch structure normally has one shared backbone with multiple branches to learn the multi-scale features. For example, Zhang~\etal~\cite{zhang2018wacv} extracted features from multiple branches and combined them for the final density estimation. Cao~\etal~\cite{cao2018eccv} improved multi-scale representations by stacking multiple scale aggregation modules, each of which consists of four branches with different kernel sizes.
Mo~\etal\cite{mo2022tip} proposed a bi-transformer that distinguishes the advantages of multi-scale branches using global attention maps and recombines multi-scale predictions following a Wenn diagram.} 

%The early version, MCNN~\cite{zhang2016cvpr}, employs the multi-column architecture to estimate density maps. %with filters of different receptive fields. 
%It is improved in subsequent works: 
%Cao \etal\cite{cao2018eccv} proposed an encoder-decoder network which includes scale aggregation modules to extract multi-scale features. %and the latter includes transposed convolutions to generate high-resolution density maps.
Multi-scale combinations in many works are simply realized via feature concatenation/averaging. Recent works leverage image perspective~\cite{shi2019cvpr,yan2019iccv,yang2020cvpr}, attention~\cite{shi2019iccv,jiang2020cvpr}, and context~\cite{liu2019iccv} information as local proxies for the combination. For example, {Shi \etal\cite{shi2019cvpr} designed a perspective-aware neural network to generate perspective maps for combining multi-scale density maps. Jiang \etal\cite{jiang2020cvpr} utilized two neural networks to generate attention maps and density maps respectively and multiply them together.}  \miaojing{Given the observed per-pixel performance discrepancy over multi-scale predictions in crowd counting, we argue that popular proxies such as perspective and attention maps are not accurate for the multi-scale combination. 
%Instead, a finer feature steering scheme on a per-pixel basis is needed. 
To this end, we introduce the hierarchical mixture of density experts to hierarchically combine the multi-scale density predictions using learnable pixel-wise soft weight maps. }

%The multi-scale outputs in our work are combined via an MoE. It is designed on a per-pixel basis to take advantage of each scale in the pixel-level and on a hierarchical basis to enable the collaboration and competition among multiple scales. 

\subsection{Mixture of experts (MoE)}
{MoE is a supervised learning method for combining systems~\cite{jacobs1991nn,jordan1994nn}.} It decomposes a task into appropriate sub-tasks and distributes them amongst multiple experts in a model. A gating net is developed to decide whether an expert is active or inactive on a per-example basis. 
%The gating decisions may be of soft or hard weights. The expert types can be versatile as stated in Sec.~\ref{sec:intro}.
In the deep learning context, MoE is used in many tasks, \eg machine translation~\cite{shazeer2017arxiv}, fine-grained categorization~\cite{zhang2019cvpr}, multi-task/modal learning~\cite{shi2019arxiv,ma2018kdd}, \etc In particular, \cite{shazeer2017arxiv} introduces a hierarchical MoE for machine translation with no overlap when grouping experts in the hierarchy. \cite{ma2018kdd} presents a single-level MoE structure yet utilizes multiple gating nets to combine experts with different sets of weights;  ~\cite{zhang2019cvpr} devises an attention module on each expert. 
MoE has also been adopted in crowd counting where experts predict global counts and are combined with image-level weights~\cite{kumagai2018MVA}. Unlike these works, our HMoDE employs pixel-wise soft gating nets to combine multi-scale density maps; group overlap is allowed in the hierarchy and attention is utilized on the gating net. These key attributes not only differ from any existing multi-scale based crowd counting methods, but are also prominent in the study of MoE.

There is a group of crowd counting methods focusing on the multi-expert model~\cite{liu2021iccv,sam2018cvpr,sam2017cvpr}. Despite also employing multiple density experts, these methods are fundamentally different from our MoE-based method: their aim is to design a routing net to separate the learning of each expert with different subsets of data.  When a new data arrives, the router finds the optimal expert for it. In contrast, our principal is to design a gating net to combine the learning of multiple experts over the entire set so that when a new data arrives, the gating net assigns the optimal weights to combine the multiple experts for it. There are also some other methods using the ensemble strategy~\cite{shi2018cvpr,zhang2021pami} in crowd counting, which clearly differ from the MoE-based structure.

\section{Method}
\subsection{Overview}\label{sec:overview}
Given a crowd image, we design a DNN to estimate the crowd density map of this image. Fig.~\ref{fig:overview} illustrates the overview of our method. We choose encoder-decoder as the basic architecture with skip connections applied (details in Sec.~\ref{sec:implementation}). Multi-scale density maps are obtained from {multiple density estimation heads} in the decoder 
% followed by two 1$\times$1 convolutions each 
and are denoted as $E_1$, $E_2$,..., $E_K$. They serve as experts for crowd density estimation. 
%Their outputs are upsampled to the original size of the input and optimized \wrt the ground truth. 
We introduce a two-level hierarchical mixture of density experts which enables the competition and collaboration of experts (Sec.~\ref{sec:MoE}). Two pixel-wise soft gating nets %($\mathcal G^1$ and $\mathcal G^2$) 
are correspondingly used for the two mixture of density experts (MoDE) modules in the two levels for multi-scale density combination. Both the combined density map and  maps over multiple scales are optimized 1) on the pixel-level using the conventional density estimation loss; and 2) on the region-level using a new relative local counting loss (Sec.~\ref{Sec:relative}). 
%The ground truth density map is generated by convolving Gaussian kernels at head centers in the image~\cite{zhang2016cvpr}. Head centers are the annotations provided for each training image.\textcolor{red}{(Annotations are provided at head centers)}     

\begin{figure*}[!ht]
\begin{center}
% \begin{overpic} 
% [width=\linewidth]
% {example-image-a}
% \end{overpic}
\includegraphics[width=\textwidth]{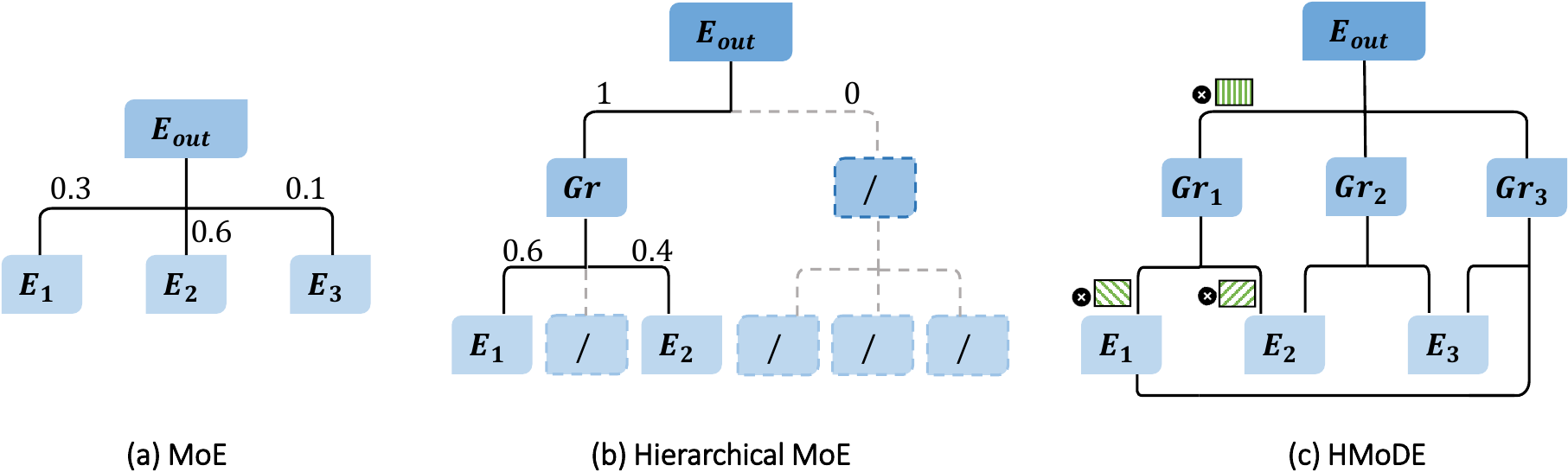}
\end{center}
\caption{\miaojing{
Illustration of HMoDE (c) and traditional MoEs (a, b). Compared with traditional MoEs, HMoDE combines multi-scale predictions in a pixel-wise manner and allows each expert to appear in more than one group.
% We show the structures of (a) MoE, (b) hierarchical MoE and (c) our HMoDE. The difference from HMoDE to traditional MoE lies in two folds: 1). the pixel-wise soft gating net. The gating net in traditional MoE generates either soft scalar weights to combine multiple experts or groups of experts with global weights (e.g. (a)); or sparse scalar weights to switch the route (e.g. (b)) to certain experts of groups of experts. The gating net in HMoDE generates pixel-wise soft weight maps to fit the dense predictions in crowd counting (see map with green lines in (c), we only show part of weight maps for simplicity). 2). the expert competition and collaboration scheme. In traditional MoE, the fusion of all experts together may lead to dominance of certain experts and suppression of other experts. To alleviate this issue, we allow group overlap so that one expert can occur in more than one group. This increases the opportunity for each expert to interact and work collaboratively with more experts, hence increases its competitivity.
}
}
\label{fig:contrihmode}
% \vspace{-5mm}
\end{figure*}

\subsection{Hierarchical mixture of density experts} \label{sec:MoE}
The original MoE uses a gating net $\mathcal G$ to distribute weights $G_1$,...,$G_K$ for the combination of multiple experts $E_1$,...,$E_K$, as illustrated in Fig.~\ref{fig:contrihmode}a. The output $E_\text{out}$ is given by $E_\text{out} = \sum_{k=1}^K G_k E_k$.   
% \begin{equation}
%     E_\text{out} = \sum_{k=1}^K G_k E_k
% \end{equation}
In order to scale up over a number of experts, a hierarchical architecture is often exploited~\cite{shazeer2017arxiv,ng2014arxiv}. It features as an inverse tree structure as illustrated in Fig.~\ref{fig:contrihmode}b: experts are partitioned into different groups and merged via weights from gating nets level-by-level from leaves to the root. Normally, each expert is assigned with one scalar weight and there is no overlap among groups at each level. \miaojing{In traditional MoE, the gating net generates either soft scalar weights to combine multiple experts or groups of experts (Fig.~\ref{fig:contrihmode}a); or sparse scalar weights to switch the route to certain experts or groups of experts~(Fig.~\ref{fig:contrihmode}b).}

Intuitively, the MoE can be used to fuse the multi-scale density outputs in a DNN for crowd counting. But it faces two issues: 1) the performance of each density expert varies by pixels on the crowd density estimation; 2) the fusion of density experts may lead to the dominance of certain experts and the suppression of other experts (this phenomenon can be reduced to a certain extent in a hierarchical MoE but still exists within each group). To cope with them, we introduce a new hierarchical mixture of density experts (HMoDE) with two highlights: the pixel-wise soft gating net and the expert competition and collaboration scheme (see Fig.~\ref{fig:contrihmode}c). We will specify them later. Below we present the basic structure of our HMoDE.   

Given $K$ experts $E_1$,...,$E_K$, we propose a two-level hierarchical MoE. In the first level, we form any $N$ ($N<K$) experts into one group. Different experts are upsampled to match the resolution of the largest one (\ie  $E_1$). Within each group, experts are fused with weights generated by the pixel-wise soft gating net $\mathcal G^1$, see Fig.~\ref{fig:overview}: HMoDE Module. Unlike the conventional hierarchical MoE, we allow group overlap such that the same expert can occur in different groups~\textcolor{blue}. The total number of groups is the combinatorial number $M =\binom{K}{N}$. Outputs of these groups are combined via another gating net $\mathcal G^2$ in the second level to produce the final output $E_\text{out}$,
\begin{equation}\label{eq:hmoe}
    E_\text{out} = \sum_{i=1}^{M}\sum_{j=1}^N G^2_i\cdot G^1_{ij} \cdot E_{ij}
\end{equation}
\miaojing{where $E_{ij}$ is the $j$-th expert in $i$-th group in the first level, $G_{ij}^1$ is the corresponding weight map generated by $\mathcal G_1$. The dot $ \cdot $ means pixel-wise multiplication. }
The number of experts $K$ in crowd counting is normally not big, the resulting computation is hence inexpensive.
%The same expert can occur in other groups such that the accumulated weight of each expert over groups is comparable to that of others. 

\subsubsection{Pixel-wise soft gating net} We design two gating nets for the two levels, respectively.  %The gating net in the first level consists of two convolution layers followed by relu; 
$\mathcal G^1$ in the first level outputs $M \times N$ weight maps $\{G^1_{ij}\}$ where every $N$ maps are softmaxed for the usage of expert combination in one group. 
$\mathcal G^2$ in the second level outputs $M$ softmaxed weight maps $\{G^2_{i}\}$. \miaojing{The softmax is applied pixel-wisely across the set of weight maps. For instance, the sum of values at the same pixel position over all weight maps in $\{G^2_{i}\}$ is one. }
Each weight map produced by $\mathcal G^1$ or $\mathcal G^2$ is pixel-wisely multiplied to the corresponding density map~{(see Fig.~\ref{fig:overview}: HMoDE Module)}. 
Both gating nets take the same feature input at the end of the encoder. We design $\mathcal G^1$ as {two} convolutional layers with each followed by a ReLU while $\mathcal G^2$ {triples} the number of layers to six. \miaojing{Notice it is rather a common practice in MoE to design the gating net as a light module~\cite{shazeer2017arxiv,chen2022nips,fedus2021jmlr,riquelme2021nips}. Despite that the gating net itself is light, its input takes the same deep encoding feature as for the experts, which contains sufficient high-level information; also, its weight assignment is guided by the network back-propagation from the high-end of the experts.}

 To distinguish the aims of the two gating nets, 
we further devise an \emph{attention module} onto the input ($F$) of $\mathcal G^2$ (see Fig.~\ref{fig:overview}: Attention Module). \miaojing{Inside this module, an attention net 
consists of {three convolutional layers followed by the Sigmoid}. It takes $F$ as the input and outputs the attention map $A$. $A$ is multiplied back to $F$ to make it focus on the foreground crowds of the image. We use the attention-guided feature $F'$ as the input to $\mathcal G^2$. The attention map is independently optimized \wrt its ground truth (see Sec.~\ref{trainingloss}: attention loss). With deep input feature and proper ground truth guidance, the attention net is capable of generating satisfactory attention. 
%This gating net with attention focuses on the fine control of experts over foreground crowds. 
During the back-propagation stage, the gradients from the foreground attended area will be reinforced through this module, resulting into finer control/optimization on this area.  } 

%a gating net as in MoE may lead to the dominance of certain experts and the suppression of other experts mixture of density experts in crowd counting. Unlike the conventional hierarchical MoE, we introduce the pixel-wise gating net for expert combination within each group, which  Considering the performance of each density expert (see Sec. 3.1 for definition) varies by pixels, we first design the pixel-wise gating nets for 

\subsubsection{Expert competition and collaboration} 
%In crowd counting, the performance of each density expert (see Sec.~\ref{sec:overview} for definition) varies by pixels; directly fusing them into one output using a gating net as in MoE may lead to the dominance of certain experts and the suppression of other experts. This phenomenon can be relieved to a certain extent in a hierarchical MoE but still exists within each group. 
\miaojing{In MoE, the expert that receives more weight during training tends to converge faster. This ends up with better performance for the expert and in turn it will receive even higher weight. The consequence is that some experts become dominant during the learning while others are suppressed. The former could lead to overfitting while the latter underfitting, which is not desired. To let all experts make effective contributions,} we look at the expert contribution on both the pixel- and image-level. Within one group, for one pixel of the density map, the estimation errors of different experts fluctuate between positive and negative values \wrt the ground truth. One expert might be regarded as less important because of its bigger error. However, if there exist other experts whose estimation errors can be statistically cancelled with that of the first expert (\ie positive and negative errors), all these experts may be regarded as important when combining them. This inspires us to allow group overlap in the HMoDE such that one expert will have more chances to interact and work collaboratively with different experts on the pixel-level {(see Fig.~\ref{fig:contrihmode}c)}. \miaojing{We give an example to illustrate this: supposing we have the {pixel-level} predictions from three experts $E_1$, $E_2$, $E_3$, with their prediction errors being -0.014, 0.004 and 0.012. If we simply merge $E_2$ with $E_1$ or $E_3$, the gating net will assign a higher weight to $E_2$ because of its smaller error, making $E_1$/$E_3$ less important in the network. On the other hand, if we allow group overlap such that $E_1$ and $E_3$ can also form a group, the error can indeed be diminished by averaging their predictions. In this sense, both $E_1$ and $E_3$ will receive balanced weights from the gating net and will make competitive contributions compared to $E_2$.}

Besides group overlap that works on pixel-level expert contributions, we also balance expert contributions on the image-level. For different experts within one group, despite their varying performance over pixels on the density estimation, we request their overall contributions on the image-level to be comparable. For each expert, we sum the pixel values of its corresponding weight map into a scalar value as a surrogate of its contribution. We regard this as each expert's importance score and introduce an expert importance loss in Sec.~\ref{trainingloss} to balance expert contributions within each group.~\miaojing{This additional constraint will enforce the inferior expert in a group to also make effective contribution. For instance, in a group of two experts, despite the overall performance of one expert may be inferior to that of the other expert at the early training stage, this loss will encourage the {inferior expert} to learn to become more competitive in the later training stage.}

% Given $K$ experts $E_1$,...,$E_K$, we propose a two-level {hierarchical} MoE. In the first level, we form any $N$ ($N<K$) experts into one group and allow group overlaps. The total number of groups is the combinatorial number $M =\binom{K}{N}$. In the first level, experts in each group are fused with weights generated by the pixel-wise gating net $\mathcal G^1$ (specified later). Group outputs are combined via another gating net $\mathcal G^2$ in the second level to produce the final output $E_\text{out}$,
% \begin{equation}\label{eq:hmoe}
%     E_\text{out} = \sum_{i=1}^{M}\sum_{j=1}^N G^2_i\cdot G^1_{ij} \cdot E_{ij}.
% \end{equation}
% $E_{ij}$ is the $j$-th expert in $i$-th group in the first level. The same expert can occur in other groups such that the accumulated weight of each expert over groups is comparable to that of others. Experts are upsampled and merged on the resolution of the largest one within each group. The number of experts $K$ in crowd counting normally falls within 3$\sim$5, the computation is totally bearable. 

%To further encourage the contribution of each expert, we introduce a pixel-oriented importance loss which is inspired by the importance loss in~\cite{shazeer2017arxiv}.(see Sec.~\ref{trainingloss}) The pixel-oriented importance loss specializes in facilitating the pixel-wise soft gating net to be more fair and generate more balanced weight maps.

\subsection{Learning from relative local counting} \label{Sec:relative}
Local count estimation used in recent works has appeared as a more appropriate target than the density map estimation \wrt evaluation criteria~\cite{liu2020eccv,xiong2019iccv,wang2021iccv}. These two means of estimation optimize crowd counting on different levels. They can be potentially complementary but can also result in conflicts in certain areas. To utilize both, we propose a new relative local counting loss alongside the absolute error loss on the density map such that they contribute to their respective targets. Measuring the relative counts among local regions can be more suitable than among pixels considering the former's robustness to noise.    

We obtain a local counting map $X$ directly from the predicted density map by integrating density values over evenly partitioned $w\times w$ regions without overlaps. The corresponding ground truth $X^g$ can be obtained in the same way on the ground truth density map. 
%We denote them by $X$ and $X^g$, respectively.
\miaojing{$X$ can also be estimated independently aside {from} the density map,  
%but it offers no additional benefits in our experiment. 
but potential conflicts may arise when optimizing the crowd density map and the local counting map together. For instance, given the same region of the image, the predicted local count from the local counting map can be higher than the ground truth local count, while the accumulated local count from the crowd density map can be lower than the ground truth. This will result into two opposite optimizing forces, hence hinders the training process. }

{Having the local counting map, we also need to design a new learning target to optimize it from a different perspective. 
% than optimizing the crowd density map
We thereby introduce the relative local counting loss, which exploits the ranking relations based on relative local counts}.
Relative local counts can be computed between any two local regions in $X$, yet easily-predicted regions contribute trivially to the optimization. To maximize efficiency, we focus on hard-predicted regions. In order to obtain these regions, we associate a local error map $R$ with $X$. 
$R$ is obtained by first calculating the pixel-wise mean squared errors between the predicted density map and the ground truth density map, then applying the same integration strategy as the local counting map $X$.  

Having $X$, $X^g$ and $R$, we select $S$ hard-predicted regions with the $S$ largest values in $R$. 
% (with the largest values in $R$). 
These regions are ranked in the descending order according to their ground truth crowd counts in $X^g$. Using this ranking order, we find the corresponding predicted crowd count of each region from $X$ and write them as a list $LX = \{x_1,x_2,...,x_S\}$. Our idea is that, 
for any two adjacent values $x_j$ and $x_{j+1}$ in $LX$, $x_j$ with a higher rank should ideally be greater than or equal to $x_{j+1}$, as this complies with their relative order indicated by their ground truth crowd counts. We can define a loss function to punish the network as long as it predicts $x_j$ to be smaller than $x_{j+1}$. 
% (Notwithstanding, the relative local count between two adjacent values in $LX$ might be disturbed by noise.) 
Nonetheless, the ground truth corresponding to $x_j$ and $x_{j+1}$ might be of very small difference in practice. A small error in the prediction can cause the inconsistency between the order of $x_j$ and $x_{j+1}$ and the order of their corresponding ground truth. This insignificant error shall trigger the loss, which is indeed undesired and may lead to the network over-fitting. Therefore, we 
% (compute the relative local count between elements of an interval of three in the list.) 
propose to split $LX$ into three sub-lists such that any two adjacent elements in a sub-list have an interval of three in $LX$:  $LX_1 = \{x_1, x_4,..., x_{s-2}\}$, $LX_2 = \{x_2, x_5,...,x_{s-1}\}$, and $LX_3 =\{x_3, x_6,...,x_{s}\}$. The sub-lists might have different sizes depending on $S$. 
Below, we define the relative local counting loss which is independently computed in each sub-list and then added together: 
\begin{equation}\label{eq:relative}
    L_\text{Rel} = \sum_{i=1}^3 \sum_{x_{j} \in LX_{i}}\max(0, x_{j+3}-x_{j}).
\end{equation}
If $x_{j+3} - x_{j}$ is smaller than zero, it is in agreement with the ground truth order and the corresponding loss is zero; otherwise, the loss occurs. The ground truth difference between two adjacent elements in a sub-list is larger than that between two adjacent elements in the original list. This makes the relative local counting loss more tolerable to insignificant errors in the prediction.

\subsection{Training loss}
\label{trainingloss}
Each density output $E$ in the network is optimized with both the relative local counting loss $L_\text{Rel}$ and the commonly used pixel-wise MSE loss $L_\text{Des}$~\cite{zhang2016cvpr,li2018cvpr}: 
%and relative distance loss are combined together to form our relative loss. The relative loss could be written as:
\begin{equation}\label{eq:des}
%\begin{split}
L_\text{Des} = \frac{1}{Z}\sum_{z=1}(e_z-e^g_z)^2
  % L_E = L_\text{Des} + {L_\text{Rel}}
  %~~~ L_\text{MSE} & = \frac{1}{2Z}\sum\limits_{z=1}^Z (e_z - e_z^g)^2 + \frac{1}{2Z_H} \sum_{z \in \mathcal Z_H} (e_z - e_z^g)^2
%  \end{split}
\end{equation}

\begin{equation}\label{eq:mse}
%\begin{split}
L_E = {L_\text{Des}} + {L_\text{Rel}}
  % L_E = L_\text{Des} + {L_\text{Rel}}
  %~~~ L_\text{MSE} & = \frac{1}{2Z}\sum\limits_{z=1}^Z (e_z - e_z^g)^2 + \frac{1}{2Z_H} \sum_{z \in \mathcal Z_H} (e_z - e_z^g)^2
%  \end{split}
\end{equation}
{where $e_z$ and $e_z^g$ are values corresponding to the $z$-th pixel in estimated density map $E$ and ground truth density map $E^g$, respectively.} Given a training image, annotations are provided at head centers in it. Its ground truth density map is generated by convolving Gaussian kernels at head centers in the image~\cite{zhang2016cvpr}. $L_\text{Rel}$ is formulated in (\ref{eq:relative}). 
%For $L_\text{MSE}$, $e_z$ and $e_z^g$ are pixel values corresponding in the estimated density map $E$ and ground truth $E^g$ respectively. $Z$ is the total number of pixels in one image. 
%For those pixels that belong to the selected hard-predicted regions in Sec.~\ref{Sec:relative} (\ie $z \in \mathcal Z_H$, $|\mathcal Z_H| = Z_H$), we double their loss values to reinforce the learning on them. 
%$\lambda$ is the loss weight. 
$L_E$ is applied on both the final density output and the intermediate density outputs in the hierarchical structure. They are all optimized on the original resolution of the image. 

Besides $L_E$, below we present the details of the attention loss and expert importance loss mentioned in Sec.~\ref{sec:MoE}.

\subsubsection{Attention loss} This loss is applied to the attention module for the second gating net $\mathcal G^2$ (see Fig.~\ref{fig:overview}). Given the predicted attention map $A$ and its ground truth $A^g$, we define the cross entropy loss between them:
\begin{equation}
\label{eq-att}
L_\text{Att}=  \frac{1}{Z}\sum_{z=1} a_z^{g}\log(a_z)-(1-a_z^{g})\log(1-a_z)
\end{equation}
where $a_z$ and $a_z^g$ are values corresponding to the $z$-th pixel in $A$ and $A^g$, respectively. \miaojing{$A$ is obtained as the output of the attention net in the attention module (Sec.~\ref{sec:MoE}-2)}; $A^g$ is obtained as the binarised result of the ground truth density map using a threshold of 1e-5~\cite{rong2021wacv}. 

\subsubsection{Expert importance loss} This loss is to encourage the balanced contribution from each expert within a group. It is applied in the first level of HMoDE. \miaojing{Given the expert $E_{ij}$ in (\ref{eq:hmoe}), we sum the pixel values of its corresponding weight map $G^1_{ij}$ to obtain an importance score $W_{ij}$ for this $j$-th expert in the $i$-th group.  We use $W_i$ to signify the set of importance scores corresponding to all experts in the $i$-th group.}
Inspired by~\cite{shazeer2017arxiv}, we compute the {standard deviation} ($\sigma(W_i)$) and mean ($\overline{W_i}$) of $W_{ij}$ in each group and formulate the expert importance loss as the squared coefficient of variation:    
%in~\cite{shazeer2017arxiv} strengthens the importance of each expert in MoE. We introduce the importance loss to our framework. In each group in the first level, we integrate the weight maps generated by the gating network into global weights $G^{Global}_{i}$. $G^{Global}_{i}=\{\sum_{z=1}^{Z_G}g^{ij}_z\}$,where $Z_G$ is the total number of pixels in a weight map $G_{ij}$ and $g^{ij}_z$ refers to the pixel values in the weight map. The variance and mean of the global weights in each group are used to calculate the coefficient of variation. At last, this loss is computed as the squared coefficient of variation.
\begin{equation}
    % L_\text{Imp}=\sum_{i=1}^M\frac{\frac{1}{N} \sum_{j=1}^N(W_{ij}-\overline{W_i})^2}{\frac{1}{N} \}
     L_\text{Eim}=\sum_{i=1}^M(\frac{\sigma(W_i)}{\overline{W_i}})^2
\end{equation}
We minimize $\sigma(W_i)$ in each group to balance the importance score of each expert on the image-level. $\sigma(W_i)/\overline{W_i}$ is a normalized form to cast $\sigma(W_i)$ from different groups into the same scale. The low standard deviation $\sigma(W_i)$ means all $W_{ij}$ are around the mean $\overline{W_i}$, \ie experts within this group are assigned with similar weights on the image level.

The total loss is therefore summarized as:
\begin{equation}
L =\sum_k L_{E_k} +  L_\text{Att}  +  L_\text{Eim}
\end{equation}
There are in total $K + M + 1$ density outputs. 

\section{Experiments}
\subsection{Datasets}

We perform our experiments on five datasets, ShanghaiTech~\cite{zhang2016cvpr}, UCF\_CC\_50~\cite{idrees2013cvpr}, JHU-CROWD++\cite{sindagi2020jhu}, NWPU-Crowd~\cite{wang2020pami}, Trancos~\cite{guerrero2015icpria} and {ShanghaiTechRGBD~\cite{lian2019cvpr}}. {ShanghaiTech} consists of two parts, SHA and SHB. Crowds in SHA are much denser than in SHB, with the average count per image being 501 and 123, respectively.  
{UCF\_CC\_50} has 50 gray images and 63974 instances. The dataset is very challenging due to its small size and large variation of images inside. To report on this dataset, hereby known as UCF, the common practice is to perform 5-fold cross validation and calculate the average result. JHU-CROWD++ is a large-scale and challenging dataset which contains 2272 training images, 500 validation images, and 1600 test images. For simplicity, we call it JHU.
NWPU-Crowd consists of 3109 training images, 500 validation images, and 1500 test images. We call it NWPU hereinafter. The results on test images are uploaded and evaluated on the NWPU CrowdBenchmark\footnote{\url{https://www.crowdbenchmark.com/}}. 
% UCF-QNRF consists of 1201 training images and 334 test images. We call it QNRF hereinafter.
{Trancos} is a vehicle counting dataset which includes 1244 images with 41976 vehicles annotations. 
{ShanghaiTechRGBD is a large-scale dataset containing 2193 images and 144512 persons. This dataset also provides the depth map for each image, but we did not use it.  We dub this dataset as RGBD for convenience.}

\begin{table*}[!t]
%\scriptsize
	%\setlength{\tabcolsep}{3.5pt}
\centering
	       % \captionsetup{font={small}}
	\caption{Comparison with state-of-the-art methods on SHA, SHB, UCF, JHU and NWPU. Ours is HMoDE + REL. The best and second best results are marked in red and blue, respectively.}
    % \resizebox{\linewidth}{!}{
	\begin{tabular}{c|c|cc|cc|cc|cc|cc}

		\toprule
		Dataset  &  & \multicolumn{2}{c}{SHA}  & \multicolumn{2}{c}{SHB} &  \multicolumn{2}{c}{UCF} & \multicolumn{2}{c}{JHU} &  
    	\multicolumn{2}{c}{NWPU}\\
		
		\midrule
		% \cline{2-7}
		Method & Venue & MAE & MSE & MAE & MSE & MAE & MSE & MAE & MSE& MAE & MSE\\
	\midrule
	    CSRNet ~\cite{li2018cvpr} & CVPR'18 & 68.2 & 115.0 & 10.6 & 16.0 & 266.1 & 397.5 & 85.9 & 309.2 & 121.3 & 387.8\\
	    SANet ~\cite{cao2018eccv} & ECCV'18 & 67.0 & 104.5 & 8.4 & 13.6 & 258.4 &334.9 & 91.1 &320.4  & 190.6 & 491.4\\
	    S-DCNet ~\cite{xiong2019iccv} & ICCV'19 & 58.3 & 95.0 & 6.7 & 10.7 & 204.2 & 301.3 &-- &-- & 90.2 & 370.5\\
	    DSSINet ~\cite{liu2019iccv} & ICCV'19 & 60.6 & 96.0 & 6.8 & 10.3 & 216.9 & 302.4 & 133.5 & 416.5 &  -- & --\\
	    PaDNet ~\cite{tian2019tip} & TIP'19 & 59.2 & 98.1 & 8.1 & 12.2 & 185.8 & 278.3 & -- & --  & -- & --\\
	    ASNet ~\cite{jiang2020cvpr} & CVPR'20 & 57.8 & 90.1 & -- & -- & 174.5& \textcolor{blue}{251.6} &-- &-- &  -- & --\\
%	    ADSCNet ~\cite{bai2020cvpr} & CVPR'20 & 55.4 & 97.7 & 6.4 & 11.3 & 198.4 & 267.3 & --&--  & \textcolor{red}{71.3} & \textcolor{red}{132.5}\\
	    AMRNet ~\cite{liu2020eccv}& ECCV'20 & 61.5 & 98.3 & 7.0 & 11.0 & 184.0 & 265.8 &-- &-- &  -- & --\\
	    DM-Count ~\cite{wang2020nips} & NeurIPS'20 & 59.7 & 95.7 & 7.4 & 11.8 & 211.0 & 291.5 & -- & -- &  88.4 & 388.6\\
%	    DensityCNN-H ~\cite{jiang2020tmm} & TMM'20 & 63.0 & 106.3 & 9.1 & 16.3& 244.5& 341.7& 3.1\\
	    LSC-CNN ~\cite{sam2020pami} & PAMI'20 & 66.4 & 117.0 & 8.1 & 12.7 & 225.6 & 302.7 & 112.7 & 454.4&  -- & --\\
	    CG-DRCN-CC~\cite{sindagi2020jhu} & PAMI'20& 60.2 &94.0 &7.5 & 12.1& -- & -- & 71.0 & 278.6 & -- & --\\

	    CRNet~\cite{liu2020tip} & TIP'20 & 56.4 & 90.4 & 7.4 & 11.9 & 203.3 & 263.4 & -- & -- &  -- & --\\
	 %   PENet & TMM'21 & 53.8 & 89.2 & 6.5 & 10.7 & 205.8 & 289.3 & 3.0\\
	   GLoss ~\cite{wan2021cvpr} & CVPR'21 & 61.3 & 95.4 & 7.3 & 11.7 & -- & -- & 59.9 & 259.5 &  79.3 & 346.1\\
	    DKPNet ~\cite{chen2021iccv} & ICCV'21 & 55.6 & 91.0 & 6.6 & 10.9 & -- & -- & --&-- &  \textcolor{blue}{74.5} & \textcolor{blue}{327.4}\\
	    UEPNet ~\cite{wang2021iccv} & ICCV'21 & \textcolor{blue}{54.6} & 91.1 & \textcolor{blue}{6.4} & 10.9 & \textcolor{blue}{165.2} & 275.9 & --& --&   -- & --\\
	    MFDC-18 ~\cite{liu2021iccv} & ICCV'21 & 55.4 & 91.3 & 6.9 & 10.3 & -- & -- & 58.1& \textcolor{blue}{221.9} &  74.7 & \textcolor{red}{267.9}\\
	    D2CNet~\cite{cheng2021tip} & TIP'21& 59.6 & 100.7 & 6.7 & 10.7 & 221.5 & 300.7& -- & -- & -- & -- \\
        ChfL ~\cite{shu2022cvpr} & CVPR'22 & 57.5 & 94.3 & 6.9 & 11.0 & -- & -- & \textcolor{blue}{57.0} & 235.7 &  76.8 &343.0\\
        {GauNet ~\cite{cheng2022cvpr}} & CVPR'22 & 54.8 & \textcolor{blue}{89.1} & \textcolor{red}{6.2} & \textcolor{blue}{9.9} & 186.3 & 256.5 & 58.2 & 245.1 & -- & -- \\
       {S-DCNet~(dcreg)}~\cite{xiong2022eccv} & ECCV'22 & 59.8 & 100.0 & 6.8 & 11.5 & -- & -- & 62.1 & 268.9 & -- & --\\
	\midrule
		%	M = 5    & 126.9 $\pm$ 10.2 & 126.9 $\pm$ 10.2 & 19.2$\pm$ 7.2  & 19.2 $\pm$ 7.2   \\
		%	\hline
		%	M= 10  &  108.6 $\pm$ 6.4 & 121.2 $\pm$ 8.3 & 17.5 & 19.2 \\
		Baseline  & -- &  66.3 & 108.4  & 7.6 & 12.7 & 234.5 & 310.3 & 63.4 & 241.5  &  81.0 & 376.7\\
		%	M = 15  & 98.66 $\pm$ 4.0 & 108.3$\pm$ 7.4 & 15.2 & 19.3 \\
		%	\hline
%		MoE   & --& 60.2 & 100.3  & --  & --  & -- & -- & --\\
		HMoDE  & -- & 56.8 & 96.5  & 6.7 & 11.1 & 177.2 & 229.4 & 58.1 &  227.5 &  77.2 & 350.4\\

		HMoDE+REL(Ours)
		& -- & \textcolor{red}{54.4} & \textcolor{red}{87.4}  &\textcolor{red}{6.2} & \textcolor{red}{9.8} & \textcolor{red}{159.6} & \textcolor{red}{211.2} & \textcolor{red}{55.7} & \textcolor{red}{214.6} &  \textcolor{red}{73.4} & 331.8\\
		\bottomrule
	\end{tabular}
    % }

	%Times are reported in seconds per image and measured on ShanghaiTech PartB.  }
	\label{tab:comparison}	
% \vspace{-5mm}
	%    \posttabspace
\end{table*}

\subsection{Implementation Details}\label{sec:implementation}

\subsubsection{Network architecture} We build an encoder-decoder architecture: the first 13 convolutional layers of the {VGG-16\_BN} (pretrained on ImageNet){~\cite{simonyan2015iclr}} are chosen as the encoder. The decoder is composed of three blocks, each with two 3$\times$3 conv + ReLU layers.  Skip connections are applied from conv4\_3 and conv3\_3 {(before the pooling layer)} of the encoder to the input of the second and third blocks of the decoder, respectively. Features are concatenated in skip connections. The density map is 
estimated from the output of each decoding block followed by a density regression head consisting of two additional 1$\times$1 conv + ReLU layers and one upsampling layer 
(there is no upsampling layer after the last decoding block).
This produces three density maps ($K =3$) serving as the density experts. We group any two experts ($N = 2$) in the first level of HMoDE.   
The details of gating nets and the attention module are specified in Sec~\ref{sec:MoE}.

\subsubsection{Training details} We augment the training set using horizontal flipping and random cropping. 
We randomly crop patches from images with a fixed size of $256\times 256$. For SHA with smaller image resolutions, we reduce the cropping size to $128\times 128$. Our model is trained for 200 epochs. We use the Adam optimizer;
% set a batch size of 4. 
the learning rate is set to $2\times10^{-5}$ and halved after 100 epochs. By default, $N$ in (\ref{eq:hmoe}) is set to 2, the size of local region ($w \times w$) in the relative local counting is set to {$\frac{1}{8}$ of the input size, and $S$ is set to 9.}
% $16 \times 16$. 
% When the input resolution is $320\times320$,  $\lambda$ in (\ref{eq:mse}) is set to 5 and $S$ for the relative local counting is set to 9; when it is $448\times448$, $\lambda$ is 2 and $S$ is 21.  
{All parameters are tuned on a single dataset and used for all experiments.}
% All parameters are tuned on validation sets. For those datasets that do not specify validation set originally, we follow {\cite{sindagi2020jhu}} to take 10\% of the training data to form a validation set.    

\subsubsection{Baseline}  We devise a baseline to show the improvements by adding our proposed components to it. It shares the aforementioned encoder-decoder architecture but directly averages three scale density maps to produce the final output.

\begin{table}[!tb]

	\centering
        % \scriptsize
    \setlength{\tabcolsep}{4pt}
	\caption{{Comparison between ours and AGCCM~\cite{mo2022tip} on SHA, SHB, JHU and RGBD.} }
	\begin{tabular}{c|cc|cc|cc|cc}
		\toprule
		Dataset  & \multicolumn{2}{c}{SHA} & \multicolumn{2}{c}{SHB} & \multicolumn{2}{c}{JHU} & \multicolumn{2}{c}{RGBD}\\
		\midrule
		% \cline{2-7}
		Method & MAE & MSE & MAE & MSE & MAE & MSE & MAE & MSE\\
	\midrule
%	Hydra 2s & 11.0 & 13.7 & 16.7 & 19.3\\
	AGCCM & 52.7 & 85.0 & 5.9 & 9.7 & 58.5 & 215.0 & 3.9 & 5.8\\

	\midrule
		Ours  &{54.4} &{87.4} & {6.2} & {9.8} & 55.7 & 214.6 & 3.7 & 5.7\\
        Ours* & 51.3 & 84.8 & 6.2 & 9.8 & -- & -- & -- & --\\
		\bottomrule
	\end{tabular}

	%Times are reported in seconds per image and measured on ShanghaiTech PartB.  }
	\label{tab:agccm}	
\end{table}

\begin{table*}[!t]
% \scriptsize

  \begin{minipage}{0.5\textwidth}
% \scriptsize
% \setlength{\tabcolsep}{3.3pt}
%\setlength{\tabcolsep}{3.5pt}
	\centering
	\caption{{Comparison with state of the arts on Trancos.} }
	\begin{tabular}{c|c|c|c|c}
		\toprule
		Dataset  & \multicolumn{4}{c}{Trancos} \\
		\midrule
		% \cline{2-7}
		Method & G0 & G1 & G2 & G3\\
	\midrule
%	Hydra 2s & 11.0 & 13.7 & 16.7 & 19.3\\
	CSRNet \cite{li2018cvpr} & 3.6 & 5.5 & 8.6 & 15.0\\
%	LC-PSPNet & 3.6 & 5.0 & 7.4& 11.7 \\
	PSDNN \cite{liushi2019cvpr} & 4.8 & 5.4 & 6.7 & 8.4\\
	S-DCNet \cite{xiong2019iccv} & \textcolor{blue}{2.9} & 4.3 & \textcolor{blue}{5.5} & \textcolor{blue}{7.1} \\
	LSC-CNN \cite{sam2020pami} & 4.6 & 5.4 & 6.9 & 8.3 \\
%	ADSCNet ~\cite{bai2020cvpr} & \textcolor{blue}{2.6} & -- & -- & -- \\
	DensityCNN \cite{jiang2020tmm} & 3.2 & 4.8 & 6.3 & 8.3 \\
	PENet \cite{yan2021tmm} & 3.1 & \textcolor{blue}{4.2} & 6.2 & 9.9 \\
% 	GauNet~\cite{cheng2022cvpr} & 2.1 & 2.6 & \\
	\midrule
		Ours  &\textcolor{red}{1.9} &\textcolor{red}{3.2} & \textcolor{red}{4.5} & \textcolor{red}{6.6}\\
		\bottomrule
	\end{tabular}

	%Times are reported in seconds per image and measured on ShanghaiTech PartB.  }
	\label{tab:comparisonTran}	
  \end{minipage}
\hfill
\noindent
\begin{minipage}{0.5\textwidth}

\renewcommand\arraystretch{1}

	\centering

	\caption{{Ablation study of the proposed HMoDE on SHA and SHB.} }
	\begin{tabular}{c|cc|cc}
		\toprule
		Dataset  & \multicolumn{2}{c}{SHA} & \multicolumn{2}{c}{SHB}\\
		
		\midrule
		% \cline{2-7}
		Method & MAE & MSE & MAE & MSE\\
	\midrule
        Average & 66.3 & 108.4 & 7.6 & 12.7\\
		Concat &  63.1 &  104.6 &7.7 &12.1 \\
	    MoE  & 61.4 & 102.3 & 7.4& 11.5\\
	    \midrule 
	     \miaojing{HMoDE-ord (($E_1$,$E_3$),$E_2$)} & 61.1 & 101.8 & 7.2 & 11.8 \\
      \miaojing{HMoDE-ord (($E_1$,$E_2$),$E_3$)} & 62.3 & 102.3 & 7.0 & 12.0 \\
      \miaojing{HMoDE-ord ($E_1$,($E_2$,$E_3$))} & 60.5 & 99.2 & 7.0 & 11.6 \\
      \midrule
	    HMoDE (w/ iw) &  60.4 &  100.8 & 7.2 & 12.0\\
	    HMoDE (w/ phw) & 59.0 & 99.2& {7.3} & 12.1 \\ 
	     HMoDE (w/ cg) & 58.9 & 100.9 & 7.0 & 11.8\\
	    HMoDE (w/o att) & 57.9 & 97.7& 6.9& 11.9 \\
	    HMoDE (w/o eim)  & 57.1 & 97.2 & 6.9& 11.6\\
	    \midrule 
	     HMoDE & \textbf{56.8} & \textbf{96.5} & \textbf{6.7} & \textbf{11.1}\\
	 %   \midrule
		\bottomrule
	\end{tabular}

	%Times are reported in seconds per image and measured on ShanghaiTech PartB.  }
	\label{tab:ablationHMoDE}	
  \end{minipage}

\end{table*}

\begin{figure*}[!ht]
\begin{center}
% \begin{overpic} 
% [width=\linewidth]
% {example-image-a}
% \end{overpic}
\includegraphics[width=0.8\textwidth]{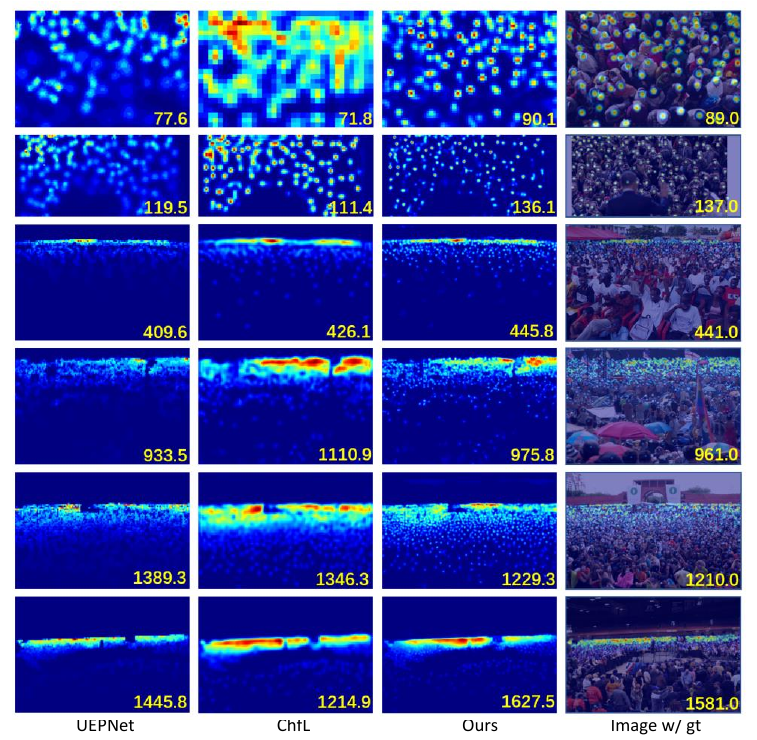}
\end{center}

\caption{
{{Qualitative results of our method compared with UEPNet~\cite{wang2021iccv} and ChfL~\cite{shu2022cvpr}.}}
}
\label{fig:results}
% \vspace{-5mm}
\end{figure*}

\subsection{Evaluation Metrics}
\label{sec:evaluation}
We use the commonly used Mean Absolute Error (MAE) and Mean Squared Error (MSE)~\cite{zhang2016cvpr} to evaluate the accuracy of our method. The Trancos dataset employs a different evaluation metric called Grid Average Mean absolute Error (GAME). It is introduced to take into account the local count accuracy, as MAE and MSE only consider the global count. Its derivation can be found in~\cite{li2018cvpr}, which has four levels, G0 $\sim$ G3. While the level increases, the number of divided regions increases and the evaluation becomes more subtle. {Results are reported on the test set of each dataset.}

\subsection{Comparisons to State of the Art}

This experiment compares our method to the state of the arts in ShanghaiTech, UCF, JHU and NWPU. The results are shown in Table~\ref{tab:comparison} where comparisons are made to recent works in crowd counting ~\cite{li2018cvpr,cao2018eccv,liu2019iccv,xiong2019iccv,jiang2020cvpr,liu2020eccv,sam2020pami,sindagi2020jhu,chen2021iccv,wang2021iccv,liu2021iccv,wan2021cvpr,wang2020nips,shu2022cvpr,cheng2021tip,liu2020tip,tian2019tip,cheng2022cvpr,xiong2022eccv}. Many of them utilize multi-scale structures~\cite{chen2021iccv,liu2019iccv,liu2020eccv,jiang2020cvpr}. Multi-scale outputs are combined via proxies such as attention and context in these works, while in our work they are combined in a hierarchical manner through pixel-wise soft gating nets. 
Local count estimation is utilized in~\cite{xiong2019iccv,liu2020eccv,wang2021iccv,xiong2022eccv}. \cite{liu2020eccv} regresses local counts based on absolute errors while \cite{xiong2019iccv,wang2021iccv} classify local counts in different intervals. {~\cite{xiong2022eccv} further loosens local count regression into a discrete ordering problem.} Density map estimation is not jointly optimized with the local count estimation in them whereas our work utilizes the relative local counting loss to enable both. Our work performs the best on these datasets over the state of the art. Sizeable differences can be observed when comparing our work to others. For instance, ours outperforms the second best on UCF by decreasing the MAE by 5.6 compared to~\cite{wang2021iccv} and the MSE by 40.4 compared to~\cite{jiang2020cvpr}. Ours also performs particularly well on the large-scale datasets, JHU and NWPU, which demonstrates its generalizability (our result on NWPU was also reported on NWPU CrowdBenchmark Leaderboard$^1$: HMoDE+REL).
%\footnote{\url{https://www.crowdbenchmark.com/nwpucrowd.html}} .   
These results justify our motivation of having per-pixel fine control on multi-scale density combination and having different optimization methods for density and local count estimations. {In Fig.~\ref{fig:results}, we show some qualitative results of our method compared with two very recent methods~\cite{wang2021iccv,shu2022cvpr}.}

\emph{AGCCM \cite{mo2022tip}.}  \miaojing{We notice a very recent work AGCCM~\cite{mo2022tip} published during the reviewing period of our work, which also uses the multi-scale architecture.
%but our designs are different: AGCCM aims to promote collaboration among multiple branches by introducing a bi-transformer to divide independent and collaboration areas for different branches while we design a hierarchical mixture of density experts to realize a finer control on the multi-scale combination. Our optimization targets are also different: AGCCM optimizes the density map while we enable the joint optimization on both density map and local counting map.}
AGCCM was originally evaluated on four datasets: SHA, SHB JHU and RGBD. In Table~\ref{tab:agccm} we report our results on these datasets:  ours works better than AGCCM on the JHU and RGBD while AGCCM is better on SHA and SHB. Nevertheless, we notice that the input patch size for AGCCM is 512 while ours is 128 for images in SHA and 256 for images in remaining datasets. We repeat the  experiment by increasing the input patch size of our method to that of AGCCM and denoting the new version as Ours*. We observe a significant improvement on SHA, \eg the MAE is decreased by 3.1 points; while the result on SHB remains the same.  We only run this experiment on SHA and SHB, \shi{as an increase of the patch size will result in significant increase of the computational complexity, which is undesirable given our hardware capacity. For instance, when the patch size is doubled from 256 to 512 for JHU, we observe a quadratic increase of GFLOPs during training.} 

}

\begin{figure}[t]
\begin{center}
% \begin{overpic} 
% [width=\linewidth]
% {example-image-a}
% \end{overpic}
\includegraphics[width=\linewidth]{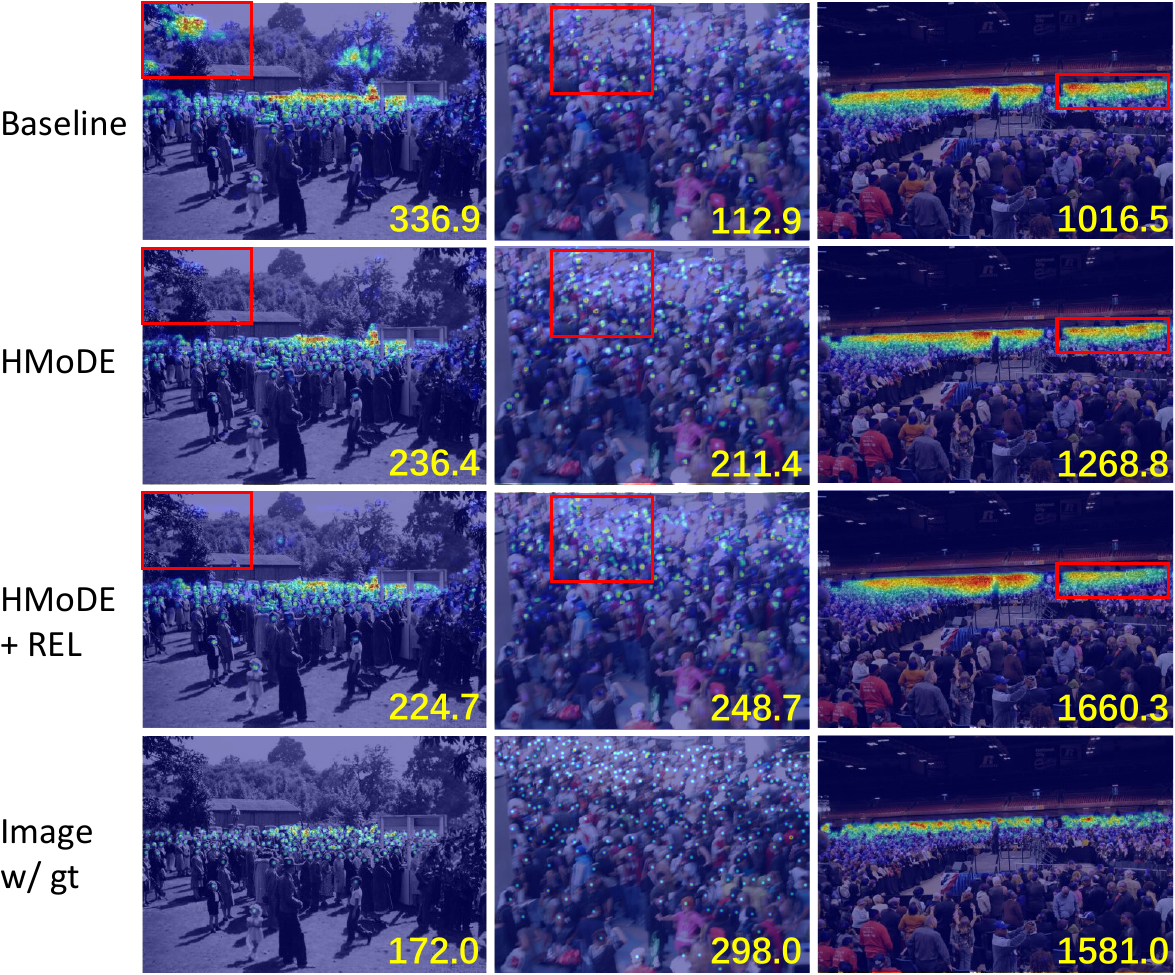}
\end{center}

\caption{
{
% Qualitative examples of ablating on proposed components of our method.
\shi{Qualitative examples of adding our proposed HMoDE and REL to the baseline.
}}
}
\label{fig:resultsrel}
% \vspace{-5mm}
\end{figure}

\emph{Trancos.}
The evaluation on Trancos is reported using GAME metrics. Results are shown in Table~\ref{tab:comparisonTran} where we compare with the state of the arts~\cite{li2018cvpr,liushi2019cvpr,sam2020pami,jiang2020tmm,yan2021tmm,xiong2019iccv}.
Our method clearly outperforms other methods over G0 $\sim$ G3, which suggests that our estimated density maps comply with the ground truth in both global counts and local details. Although~\cite{liushi2019cvpr,sam2020pami} were specifically designed for crowd localization (favoured by GAME), our method outperforms theirs. This shows the good generalizability of our method.

\subsection{Ablation Study }
\label{sec:ablation}
The ablation study for two key contributions, hierarchical mixture of density experts (HMoDE) and learning from relative local counting (REL), is provided in Table~\ref{tab:comparison} for all datasets. We can see significant improvement from the baseline by adding HMoDE and REL ($L_\text{Rel}$ in (\ref{eq:mse})). \miaojing{Some qualitative results of adding HMoDE and REL to the baseline are shown in Fig.~\ref{fig:resultsrel}.} Below we offer an ablation study on more detailed elements and conduct experiments on SHA and SHB.

\indent \emph{1) Hierarchical mixture of density experts}\\
\indent \emph{Feature fusion. } We first study the feature fusion by comparing MoE to  the concatenation and averaging operations. For simplicity, we do not use the hierarchical structure at this stage. All density maps from three scales are fused into one map via the proposed \emph{pixel-wise soft gating net} in the MoE. For feature concatenation, we concatenate the feature maps over three scales and apply a
1$\times$1 conv to obtain the final density map. For averaging of multi-scale density maps, it is our baseline (Table~\ref{tab:comparison}). The results are shown in Table~\ref{tab:ablationHMoDE}. Averaging multi-scale density outputs produces an MAE of 66.3 and an MSE of 108.4 on SHA. Using feature concatenation produces lower MAE and MSE. 
%The concatenated feature map is projected into the density map via a 1 $\times$ 1 conv, where weights are shared among spatial positions. 
Our MoE with pixel-wise weights lowers the MAE by 1.7 points and the MSE by 2.3 points on SHA compared to the variant using concatenation. 
{Similar observation can be found on SHB. }

\indent \textit{Hierarchical structure.} Following the improved performance of MoE in the single-level, the proposed HMoDE extends it to two levels, which significantly reduces the MAE to 56.8{~(6.7)} and MSE to 96.5{~(11.1)} on SHA{~(SHB)}. % exhibits that the multi-level hierarchical architecture is superior to MoE architecture. 
In HMoDE, we introduce an \emph{expert competition and collaboration} scheme which allows group overlap when mixing density experts. An ordinary hierarchical MoE normally does not have this overlap. To implement this {variant} in our setting with three experts, in the first level~\miaojing{we directly pass one expert to the second level and combine the other two with a pixel-wise gating net. 
For instance, we can combine $E_1$ and $E_3$ with a pixel-wise soft gating net and let $E_2$ directly pass to the second level; we use HMoDE-ord (($E_1$, $E_3$),$E_2$) to denote this variant and similarly other variants in Table \ref{tab:ablationHMoDE}}. Results show that, with a proper combination of $E_1$, $E_2$ and $E_3$, HMoDE-ord can also improve the performance from MoE but the improvement is much less than our HMoDE. 

\indent \textit{Pixel-wise soft gating.} The multi-scale density experts are combined via per-pixel softmaxed weights in HMoDE.  In an ordinary gating net, density experts are combined with only image-level weights. We alter our pixel-wise soft gating net to behave like an ordinary gating net and denote it as HMoDE (w/ iw). This combination is coarse and ends up with an MAE of 
60.4 and an MSE of 100.8 on SHA. Next, with respect to the soft weights, we can also alter the gating net to produce pixel-level hard weights such that only one expert is selected at each pixel. This, denoted by HMoDE (w/ phw), produces an MAE of 59.0 and MSE of 99.2 on SHA. 
% Instead of using the pixel-wise soft gating net for the proposed hierarchical mixture of density experts (HMoDE), 
Third, compared to pixel-wise soft gating, we tried an alternative version, the channel-wise soft gating. It produces channel-wise soft weights to combine multi-scale features from different decoding blocks (before projecting them to density maps). We denote this by HMoDE (w/ cg) in Table \ref{tab:ablationHMoDE}. The MAE and MSE on SHA are 58.9 and 100.9 respectively, which are inferior to pixel-wise soft gating nets. {Similar observation can be found on SHB.} 

\begin{table*}[!t]
% \scriptsize

    \begin{minipage}{0.5\textwidth}
\centering
% \vspace{-2mm}
	% \footnotesize
% \setlength{\tabcolsep}{1.5pt}

	\caption{\miaojing{Results on different designs of the gating net 1. Our default setting is with 2 conv layers. } }
	\begin{tabular}{c|cc|cc}
		\toprule
		 Dataset & \multicolumn{2}{c}{SHA} & \multicolumn{2}{c}{SHB}  \\
		
		\midrule
		% \cline{2-7}
		HMoDE-Gating net 1 & MAE & MSE & MAE & MSE\\
	\midrule

		2*Conv & 56.8  & 96.5 & 6.7 & 11.1 \\
        \midrule
	%	+ ALC(MAE) & -- & -- \\
		4*Conv & 57.3 & 101.1 & 6.7 & 11.2 \\
        6*Conv & 57.2 & 97.7 & 6.8 & 11.3\\
        ResBlock & 56.8 & 96.3 &6.7 & 11.2 \\
       MSBlock & 57.8 & 96.2 & 6.7 & 11.3 \\
	%	Ord($K$=4,$N$=2) & 55.4 & 90.4 & 6.6 & 10.8 \\
		\bottomrule
	\end{tabular}
% 	\vspace{18mm}

	%Times are reported in seconds per image and measured on ShanghaiTech PartB.  }
	\label{tab:gatestruc}	
  \end{minipage}
  \hfill
  \noindent
% \scriptsize
\begin{minipage}{0.5\textwidth}
	\centering
	\caption{\miaojing{Results of using input features from different positions of our architecture to the gating net 1.} }
	\begin{tabular}{c|cc|cc}
		\toprule
		Dataset  & \multicolumn{2}{c}{SHA} & \multicolumn{2}{c}{SHB}\\
		
		\midrule
		% \cline{2-7}
		HMoDE-Gating net 1 & MAE & MSE & MAE & MSE\\
	\midrule

		EB1 & 59.3 & 100.1 & 6.9 & 11.5\\
        EB2 & 57.0 & 96.8 & 6.8 & 11.2\\
		EB3  & \textbf{56.8} & \textbf{96.5} & \textbf{6.7}& \textbf{11.1}  \\
        DB3 & 57.1 & 97.4 & 6.9 & 11.5\\
        DB2 & 58.9 & 98.8 & 7.1 & 11.3\\
        DB1 & 59.1 & 98.9 & 7.1 & 11.6 \\
		\bottomrule
	\end{tabular}

	\label{tab:gateInput}
\end{minipage}

\end{table*}

\miaojing{Our gating net is a light module. To justify our design, we provide additional experiments for different variants of the gating net 1 by doubling/tripling the number of convolutional layers in gating net 1, as well as transforming it into a residual block (ResBlock)~\cite{he2016cvpr} or multi-scale block (MSBlock)~\cite{cao2018eccv}. The results in Table~\ref{tab:gatestruc} show that making the gating net more complex does not bring us additional benefits. 
Besides the module structure, we also justify the selection of the input feature to this {gating net} by varying the input feature position after every encoding/decoding block in our architecture (see Fig.~\ref{fig:overview}). The results in Table~\ref{tab:gateInput} show that using the feature after the encoder (EB3, our default setting) performs the best. This  feature contains rich visual and semantic information that is shared across the density experts. Features from earlier layers of the encoder might not contain sufficient semantic information for controlling the density experts, while features from the later layers of the decoder will not be shared for all experts hence are not suitable for the gating net. 
}
% In contrast, the results of our original HMoDE with pixel-wise soft gating nets are superior. 
%This justifies our design of the pixel-wise gating net.
\begin{table*}[!t]
% \scriptsize
\begin{minipage}{0.5\textwidth}
    \centering
% \vspace{-2mm}
	\footnotesize

	\caption{\miaojing{Results on more complex designs of the attention net. Our default setting is with 3 conv layers. } }
	\begin{tabular}{c|cc|cc}
		\toprule
		 Dataset & \multicolumn{2}{c}{SHA} & \multicolumn{2}{c}{SHB}  \\
		
		\midrule
		% \cline{2-7}
		HMoDE-Attention net & MAE & MSE & MAE & MSE\\
	\midrule

		3*Conv & 56.8  & 96.5 & 6.7 & 11.1 \\
        \midrule
	%	+ ALC(MAE) & -- & -- \\
		6*Conv & 57.2 & 97.9 & 6.7 & 11.4 \\
        9*Conv & 57.0 & 97.3 & 6.8 & 11.6\\
        ResBlock & 56.9 & 96.8 &6.9 & 11.9 \\
       MSBlock & 57.3 & 97.4 & 6.9 & 11.7 \\
		\bottomrule
	\end{tabular}
% 	\vspace{18mm}

	%Times are reported in seconds per image and measured on ShanghaiTech PartB.  }
	\label{tab:attstruc}	
\end{minipage}
  \hfill
  \noindent
    \begin{minipage}{0.5\textwidth}
\centering
% \vspace{-2mm}
	\footnotesize
\setlength{\tabcolsep}{5pt}

	\caption{{\miaojing{Ablation study of increasing the width and depth of our architecture. }} }
	\begin{tabular}{c|cc|cc|c|c}
		\toprule
		 Dataset & \multicolumn{2}{c}{SHA} & \multicolumn{2}{c}{SHB} & \multicolumn{1}{c}{Mem} & Time  \\
		
		\midrule
		% \cline{2-7}
		Method & MAE & MSE & MAE & MSE & GB & Sec.\\
        \midrule
        Ours($K$=3,$N$=2,$D$=2) & 54.4 & 87.4 & 6.2 & 9.8 & 12.05 & 55\\
	\midrule

		Ours($K$=4,$N$=2,$D$=2) & 55.0  & 90.8 & 6.5 & 10.5 & 14.81 & 69\\
	%	+ ALC(MAE) & -- & -- \\
		Ours($K$=4,$N$=3,$D$=2) & 54.2 & 86.9 & 6.2 & 9.6 & 15.45 & 64 \\
        Ours($K$=3,$N$=2,$D$=3) & 55.4 & 92.7& 6.8 & 11.4 & 17.61& 59\\
		\bottomrule
	\end{tabular}
% 	\vspace{18mm}

	%Times are reported in seconds per image and measured on ShanghaiTech PartB.  }
	\label{tab:ablationwider}	
  \end{minipage}

\end{table*}

We also evaluate the effect of the attention module in the second gating net in HMoDE. Table~\ref{tab:ablationHMoDE} shows that HMoDE without the attention (HMoDE w/o att) increases the MAE by {+1.1~(0.2)} and MSE by {+1.2~(0.8)} on SHA~(SHB), respectively. The attention module helps the second gating net focus more on foreground regions. To justify this purpose, we further evaluate the model on foreground regions of test images (we follow the same way as the generation of $A^g$ (\ref{eq-att}) to obtain the foregrounds): we observe that without the attention module, the performance is degraded more severely (\eg +5.8/+19.9 on MAE/MSE on SHA) on foreground regions than on whole images. This shows the particular contribution of the attention module on foreground density prediction.

\miaojing{Our attention net is lightweight, we provide some variants of it by doubling (6*Conv) or tripling (9*Conv) the number of convolution layers, or transforming it into a residual block (ResBlock)~\cite{he2016cvpr} or a multi-scale block (MSBlock)~\cite{cao2018eccv}.  Results in Table~\ref{tab:attstruc} demonstrate that making the attention net more complex does not bring us additional benefits.   }

\indent \textit{Expert importance loss.} This loss is introduced to balance the expert contributions in each group. We report the results of not using it in HMoDE (Table \ref{tab:ablationHMoDE}: HMoDE (w/o eim)). For instance, an improvement of 0.3 on MAE and 0.7 on MSE is achieved on SHA, demonstrating the impact of the expert importance loss. %\textcolor{red}{Consistent improvement can be observed on SHB.}

\indent \textit{Going wider and deeper. } By default, the number of density experts $K$ is 3, we can widen the proposed architecture by increasing $K$ to 4. In this context, the number of experts per group $N$ can be either 2 or 3. We report both results (full version: HMoDE+REL) in Table~\ref{tab:ablationwider} which shows that $K$=4,$N$=3 works better than $K$=4,$N$=2; for instance, on SHA, the MAE and MSE decrease by {0.8} and 3.9 points from the former to the latter.
\miaojing{We have also listed the occupied memory and time per epoch during training for each variant on SHA in the table.} 
% (our default setting, see Table~\ref{tab:comparison}). 
Compared to our default setting ($K$=3,$N$=2) (Table~\ref{tab:comparison}), the improvement by increasing $K$ to 4 ($K$=4,$N$=3) is insignificant, considering that the computational cost also increases in this process. %We will have to consider this tradeoff when going wider. 

Apart from going wider, we've also investigated the effect of extending our HMoDE by increasing its depth ($D$) from two levels to three levels: density experts are combined into overlapping groups in the first and second levels, and all combined in the third level.~\miaojing{We report its result in Table~\ref{tab:ablationwider}, denoted by Ours($K$=3,$N$=2,$D$=3).} It produces no further benefits and the performance is indeed degraded. We suspect this is due to that the number of density experts $K$
%comparing to dozens/hundreds of experts in~\cite{shazeer2017arxiv},  
is not big in this work. Having two-level HMoDE is sufficient to handle it while having three levels tends to over-fit. On the other hand, increasing $K$ may not be the solution, as it also depends on other factors, \eg the dataset size, whether we are allowed with the capacity of having a very big $K$.

\indent \emph{2) Learning from relative local counting} 

\begin{table}[t]

	\centering
	\caption{{Ablation study of local region losses.} }
	\begin{tabular}{c|cc|cc}
		\toprule
		Dataset  & \multicolumn{2}{c}{SHA} & \multicolumn{2}{c}{SHB}\\
		
		\midrule
		% \cline{2-7}
		Method & MAE & MSE & MAE & MSE\\
	\midrule

		HMoDE  & 56.8 & 96.5 & 6.7& 11.1  \\
		      % & 0.9 & 1.8 & 0.4 & 1.1\\
	%	+ ALC(MAE) & -- & -- \\
	
        \miaojing{+ GRL} & 56.2 & 95.4 & 6.6 & 11.1 \\
        	+ ALC & 57.0 & 97.2 & 6.7& 11.4\\
            \midrule
        		% + DB + ALC & 57.4 & 97.0 & 6.9& 11.5\\
		+ DB + REL & 57.4 & 96.2 & 7.5 & 12.0\\
	     % + OB + ALC & 57.0 & 97.2& 6.7 & 11.4 \\ 
		+ OB + REL& \textbf{54.4} & \textbf{87.4} & \textbf{6.2}& \textbf{9.8} \\
  \midrule
		+ REL (RR) & 56.1 & 95.4 & 6.6 & 10.6\\
	   + REL (MD) & 59.3 & 100.1& {8.2} & {13.5} \\ 
		+ REL (HR) & \textbf{54.4} & \textbf{87.4} & \textbf{6.2}& \textbf{9.8} 
		\\
		\bottomrule
	\end{tabular}

	%Times are reported in seconds per image and measured on ShanghaiTech PartB.  }
	\label{tab:ablationLoss}

\end{table}

% \noindent \textcolor{red}{\textit{Improvement on GAME.}~Since we use local counting map in relative local counting loss, GAME is a suitable evaluation metric. In Table~\ref{tab:gamerel}, we show the improvement on GAME (G0$\sim$G3) by applying the REL loss in SHA, which is sizeable.}
\miaojing{\indent \textit{Local region vs. global image.} 
Our learning scheme operates on the local counting map. ~\cite{yang2020eccv} directly regresses crowd numbers of images without the supervision of pedestrian locations, then rank the global counts of images resembling the ranking loss in image retrieval~\cite{zhao2015cvpr,gordo2016eccv}. We follow~\cite{yang2020eccv} to generate a ground truth ranking order matrix based on ground truth crowd counts of images; transform the ranks of crowd counts obtained from estimated density maps into a predicted ranking order matrix; then calculate the cross-entropy loss between the predicted matrix and ground truth matrix.  We denote this variant as global ranking loss (GRL) in Table~\ref{tab:ablationLoss}. 
% It can be seen that +GRL does not bring additional benefits to HMoDE.
It can be seen that +GRL only slightly improves the performance upon HMoDE, while {our scheme (+REL)} significantly improves the performance,  demonstrating that it is the right way to go from the global crowd count to the local counting map.   
}

\indent \textit{Relative error vs. absolute error.} We introduce a relative error based loss for local counting map. We have also aimed to optimize the local counting map based on absolute local count (ALC) errors, \ie MAE. Table~\ref{tab:ablationLoss} shows the result of +ALC, which has no benefits on the performance. Due to the possible conflicts that may arise from the density map estimation, the performance is slightly lower than the original HMoDE.  %The MAE and MSE of these two variants are respectively xx, xx and xx, xx.  
%This validates our design of relative local counting loss. 

\indent \textit{Single branch vs. double branches.}
\miaojing{Our local counting map that is obtained from the estimated density map, \ie using a single branch; while local counting map can also be generated with its own branch independently from the density estimation branch, \ie using double branches. In Table~\ref{tab:ablationLoss}, we present the results of using one branch (OB) or double branches (DB) for density map and local counting map estimations.
%and using 
% absolute local counting loss (ALC) or 
%\shi{relative local counting loss (REL) for local counting map optimization.} 
Our default setting is + OB + REL, which is 
% the best 
{clearly better than + DB + REL} in Table~\ref{tab:ablationLoss}. 
% All other variants 
It can be seen that + DB + REL degrades the overall performance compared to HMoDE, which supports our claim that joint optimization of density map and local counting map is not straightforward. Our strategy for the first time enables it.
% cannot be easy due to potential optimization conflicts. Our strategy \shi{for the first time} enables a joint optimization on both density map and local counting map.
}

\indent \textit{Hard regions vs. random regions.} The relative local counting loss is defined on hard-predicted local regions that have large counting errors (+REL (HR)). This improves the model robustness and reduces its computational cost. Table~\ref{tab:ablationLoss} also shows the results without using hard mining but with only randomly sampled regions (+REL (RR)). This yields the MAE of 56.1 and MSE of 95.4 on SHA {and the MAE of 6.6 and MSE of 10.6 on SHB. }
Using random regions brings moderate improvement, but is less powerful than using hard-predicted regions.

\indent \textit{Non-overlap vs. overlap.} Our hard regions are non-overlapped. However, we notice in~\cite{liu2018bcvpr,liu2019tpami} they compare the local counts of overlapped local regions for unsupervised learning. They randomly select one anchor point from the anchor region in each image; find the largest square patch centered at this anchor point; reduce the patch size by a scale factor 0.75 to obtain four smaller patches. This is like the matryoshka doll (MD). We follow this practice to devise a variant of our REL (denoted by REL (MD)) and show the results on SHA and SHB in Table \ref{tab:ablationLoss}. It can be seen that + REL (MD) did not bring benefits to HMoDE. This shows that optimizing the local count prediction following the matryoshka way would not work in the fully supervised learning.   
%both the density estimation loss and the ranking loss does not work. 

\indent \textit{Generalizability of relative local counting loss.} 
% The proposed relative local counting loss is complementary to the conventional density estimation loss and is applicable to many other methods.
To verify that the proposed relative local counting loss is applicable to other methods, we apply it to two established pipelines with trainable codes, \ie CSRNet~\cite{li2018cvpr} and {DM-Count~\cite{wang2020nips}}. The results are shown in Table \ref{tab:relother}. By applying our relative local counting loss (REL) to CSRNet and DM-Count, both their MAE and MSE are clearly decreased (\eg on SHA, -3.5 and -2.1 MAE for CSRNet and DM-Count, respectively; {and consistent decrease on SHB}). This shows the effectiveness of our REL on other methods.

\begin{table*} [!ht]
    \centering

      \begin{minipage}{0.48\textwidth}
\centering
		\caption{{Relative local counting.} }
	\begin{tabular}{c|cc|cc}
		\toprule
		Dataset  & \multicolumn{2}{c}{SHA} & \multicolumn{2}{c}{SHB}\\
		
		\midrule
		% \cline{2-7}
		Method & MAE & MSE & MAE & MSE\\
	\midrule
		CSRNet~\cite{li2018cvpr} & 68.2 & 115.0 & 10.6 & 16.0\\
		+REL & 64.7 & 107.9 & 8.6 & 13.2\\
		\midrule
		DM-Count~\cite{wang2020nips} & 59.7 & 95.7 & 7.4 & 11.8\\
		+REL & 57.6 & 95.4 & 7.1 & 10.9\\
		\bottomrule
	\end{tabular}

    % \captionsetup{font={scriptsize}}

	%Times are reported in seconds per image and measured on ShanghaiTech PartB.  }
	\label{tab:relother}	
  \end{minipage}
  \hfill
  \noindent
    \begin{minipage}{0.48\textwidth}

  \centering
  	\caption{{Ablation study on the number of sub-lists in REL.} }
		\begin{tabular}{c|cc|cc}
		\toprule
		Dataset & \multicolumn{2}{c}{SHA} & \multicolumn{2}{c}{SHB} \\
		\midrule
		% \cline{2-7}
		Method & MAE & MSE &MAE & MSE \\
	\midrule

        REL(1s) & 57.2 & 97.1 &  7.0 & 11.9 \\
        REL(2s) & 55.7 & 94.4 &  6.8  & 11.0   \\
        REL(3s) &\textbf{54.4} & \textbf{87.4} & \textbf{6.2} & \textbf{9.8}     \\
        REL(4s) &56.1 & 94.1 &  6.6  & 10.5   \\
        REL(5s) &56.3 & 95.0 &  6.6  &  10.8  \\

		\bottomrule
	\end{tabular}

	%Times are reported in seconds per image and measured on ShanghaiTech PartB.  }
	\label{tab:ablationsublist}	
    \end{minipage}

\end{table*}
\begin{table*}[!ht]

          \centering
    % \renewcommand{\arraystretch}{0.75}
    % \scriptsize
       \caption{Performance of experts, groups, and final output on JHU.}
    \begin{tabular}{c|cc|cc|cc|cc|cc|cc|cc}
    \toprule
       {JHU} &  \multicolumn{2}{c}{\emph{E$_1$}} &  \multicolumn{2}{c}{\emph{E$_2$}} &  \multicolumn{2}{c}{\emph{E$_3$}} &  \multicolumn{2}{c}{$Gr_1$} &  \multicolumn{2}{c}{$Gr_2$} &  \multicolumn{2}{c}{$Gr_3$} &  \multicolumn{2}{c}{$E_\text{out}$}\\
    \midrule
        Density Level& MAE & MSE & MAE & MSE & MAE & MSE & MAE & MSE & MAE & MSE & MAE & MSE & MAE & MSE\\
    \midrule
        Overall & 58.1 & {237.8} & 58.8 & 239.0& 59.5& 242.4 &{57.0} &227.3 &57.3 & 227.0&57.6 & 229.3& 55.7 & 214.6\\
        Low & 9.8& 24.1 &9.5 & 24.8&10.3 & 27.2& 8.6& 23.1& 8.8& 23.7& 8.8& 22.8& 8.3 & 21.9\\
        Medium & 31.5& {61.3}& 33.1& 64.6&33.0 &62.9 &30.7 &50.7 &30.8 &50.9 &{31.5} &52.0 &29.2 &48.3\\
        High & 250.5 & 604.7& 249.9 &605.5 & 253.1 & 615.0& {248.4} & 579.9& 249.6& 579.0& {248.8}& 584.8& 245.6& 547.4\\
        % MSE & & & & & & & 93.2\\
    \bottomrule
    \end{tabular}
    \label{tab:expandgroup}
    
\end{table*}

\begin{figure}[t]
	\centering

\begin{center}
% \begin{overpic} 
% [width=\linewidth]
% {example-image-a}
% \end{overpic}

\includegraphics[width=1.0\linewidth]{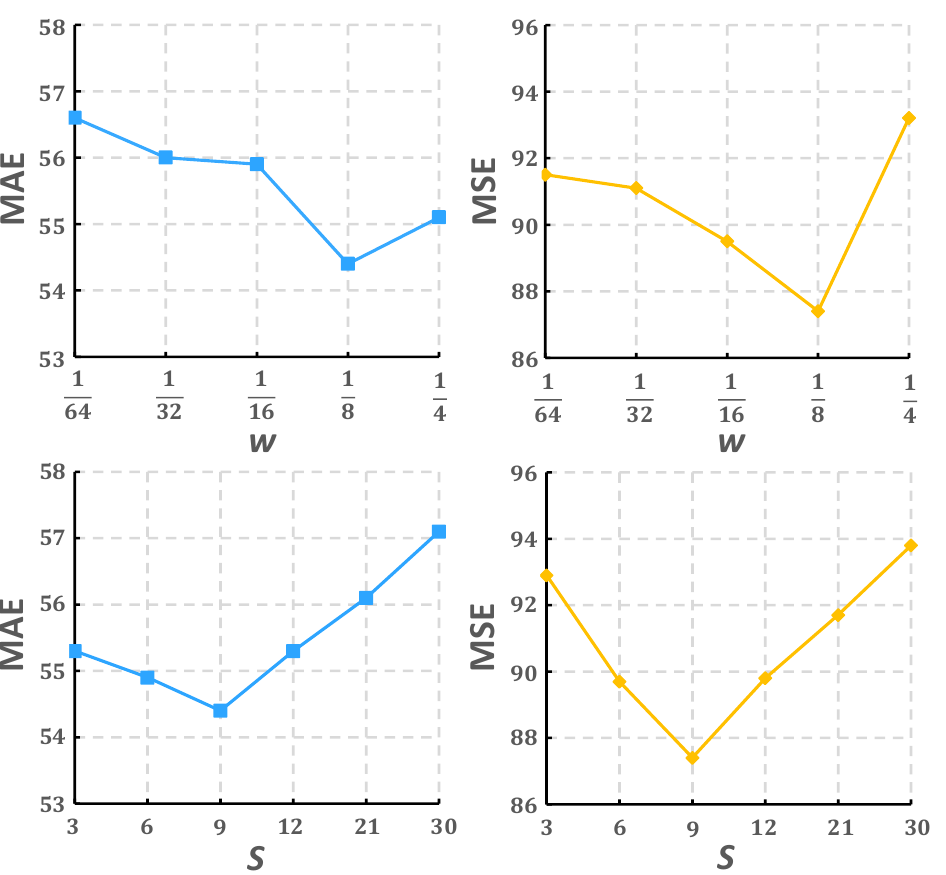}
\end{center}
% \vspace{-2mm}
\makeatletter\def\@captype{figure}\makeatother
\caption{
{Parameter variations of $w$ and $S$ on SHA. MAE and MSE are reported in the left and right.}}
\label{fig:parameter}
\end{figure}

\indent \textit{Size of local region $w\times w$.} {We vary $w$ over {$\frac{1}{64}$, 
$\frac{1}{32}$, $\frac{1}{16}$, $\frac{1}{8}$, $\frac{1}{4}$ of the input size}
% 4, 8, 16, 32, 64 
and report the MAE/MSE {on SHA} in Fig~\ref{fig:parameter}: Top. The best MAE/MSE occurs at {$w = \frac{1}{8}$ of the input resolution}, with no improvement using a larger or smaller $w$. This is our default setting.} %On the other hand, the computational cost increases proportionally to S.
%We choose $S = 21$ as the default setting for UCF.}
% \input{fig/results}

\indent \textit{Number of selected hard-predicted regions $S$.} We vary $S$ over {3, 6, 9, 12, 21, 30} and report the MAE/MSE in Fig~\ref{fig:parameter}: Bottom. The MAE/MSE decreases quickly from  $S$ being 3 to 9 and  slowly increases afterwards. %On the other hand, the computational cost increases proportionally to S.
We choose {$S = 9$} as the default setting.

\indent{\textit{Number of sub-lists in $LX$.} We perform this experiment by splitting the original list $LX$ into 1, 2, 3, 4 and 5 sub-lists and report their results in Table~\ref{tab:ablationsublist}. It shows that splitting $LX$ into 3 sub-lists (our default setting, REL(3s)) works the best.}

% \noindent \textcolor{red}{
% \textit{Comparison on the relative local counting loss.}
% We notice the ranking loss (RKL) in~\cite{liu2018bcvpr,liu2019tpami} also compares local regions by their counts. Our relative local counting loss (REL) differs from the RKL in~\cite{liu2018bcvpr,liu2019tpami} fundamentally: 1) the RKL is applied on unlabeled images without ground truth while our REL is applied on labeled images with ground truth. We for the first time enable a joint optimization on both the crowd density map and the local counting map of the same image. 2) The RKL operates on a series of square regions centered at the same location of the image with different sizes. Our REL instead operates on a series of evenly squared non-overlap regions in the image. 3) Because of no ground truth provided, RKL randomly selects one region in each image and generates a set of regions upon it to apply the ranking loss. REL however selects a number of hard-predicted regions in each image based on the ground truth and compares relative local counts among them.
% The RKL has been later employed in~\cite{cheng2019iccv}, but its basic spirit remains. The above summary of differences to ours hold primarily.
% }

\indent \emph{3) Expert analysis and visualization}

\indent In order to show the advantage of forming expert groups in our method, in this session, we analzye the intermediate results of individual experts and their formed groups and also provide qualitative results for them.

In Table~\ref{tab:expandgroup}, we report the performance of our three experts in the first level ($E_1\sim E_3$), groups in the second level ($Gr_1\sim Gr_3$) and final output E$_\text{out}$ on the JHU dataset. {According to Fig.~\ref{fig:overview}, $Gr_1$, $Gr_2$, $Gr_3$ is a weighted combination of $E_1$ and $E_2$, $E_1$ and $E_3$, $E_2$ and $E_3$, respectively. The loss in (\ref{eq:mse}) is applied to every density map so that each performs well.}
%We also give the overall results of Ex1-3, Gr1-3 and Fo on JHU in Table~\ref{tab:expandgroup}. 
% Ex1-3 denote the results of experts in the first level, Gr1-3 represent the results of groups in the second level; Fo denotes the final output. 
The predictions are getting better from experts to corresponding groups, and the final output, which demonstrates the effectiveness of our proposed hierarchical mixture of experts structure. 
\miaojing{We notice that the performance of a single expert is already quite good (\eg $E_1$). 
Two factors contribute to the competitive performance of the expert: first, instead of 
% using a vanilla VGG 
employing the vanilla VGG backbone, we follow~\cite{rong2021wacv,wang2021iccv,cheng2022cvpr} to extend the VGG backbone into an encoder-decoder with skip connections, which proves to be a much more powerful backbone; second, our proposed HMoDE and REL not only improve the performance of the final prediction, but also the intermediate predictions.}

\begin{figure*}[!ht]
\begin{center}
% \begin{overpic} 
% [width=\linewidth]
% {example-image-a}
% \end{overpic}
\includegraphics[width=\textwidth]{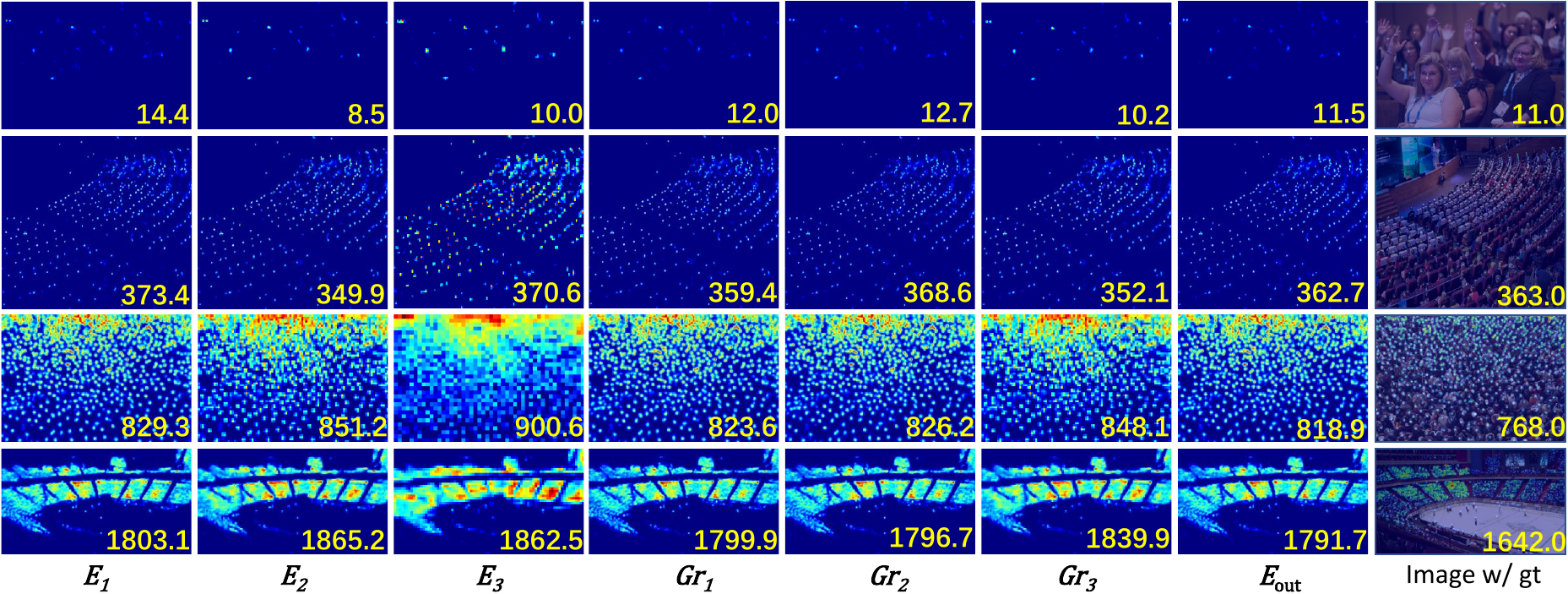}
\end{center}
\caption{{Qualitative examples of experts, groups and final output on JHU.}
}
\label{fig:jhuexp}
% \vspace{-5mm}
\end{figure*}

\begin{figure}[t]
\begin{center}
% \begin{overpic} 
% [width=\linewidth]
% {example-image-a}
% \end{overpic}
\includegraphics[width=\linewidth]{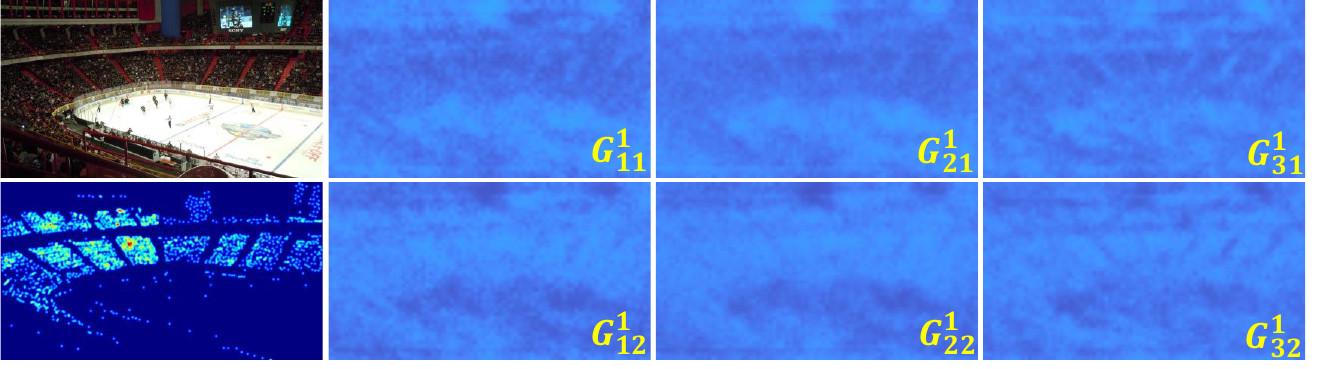}
\end{center}

\caption{
\miaojing{
Visualization of weight maps of $\{G^1_{ij}\}$ from $\mathcal G^1$. 
% The left column is an image with its ground truth density map where we sample two local regions and place them in the middle column. We run a three-scale crowd density estimation model (see Sec.~\ref{sec:implementation}: Baseline) on this image and illustrate the best performing scale in the two local regions in the right column. Blue, yellow, and green represent each scale, respectively.
{From left to right, we give the image and its ground truth density map, and weight maps for each group, respectively.}
} }
%It's obvious that one scale does not always perform well on a similar distribution.}
\label{fig:visW}

\end{figure}

Some qualitative examples are given in {Fig.~\ref{fig:jhuexp}} which are consistent with our quantitative results. For the last example, \miaojing{we also visualize its weight maps in Fig.~\ref{fig:visW}. In each weight map, lighter color indicates higher weight. One can see that weight maps in each group (column) are in general balanced so that both experts contribute to the combined prediction. This validates the efficacy of the expert competition and collaboration scheme.}

Finally, the JHU has separated its test images into three subsets according to the global crowd count in each image: \emph{Low} (0-50 people), \emph{Medium} (51-500 people) and \emph{High} (500+ people). We report results in these subsets where no significant performance difference among experts can be observed. This validates our design of \emph{expert competition and collaboration} scheme to balance contributions among experts. The observation also supports our claim that assessing the expert performance discrepancy via proxies such as crowd count distribution would not be sufficient.

% \noindent \textcolor{red}{\textit{Performance of each expert and group.} We also give the overall results of Ex1-3, Gr1-3 and Fo on JHU in Table~\ref{tab:expandgroup}. 
% % Ex1-3 denote the results of experts in the first level, Gr1-3 represent the results of groups in the second level; Fo denotes the final output. 
% The predictions are getting better from experts to groups, and the final output, which demonstrates the effectiveness of our proposed hierarchical mixture of experts structure.  }

\section{Conclusions}
In this paper, we first propose a hierarchical mixture of density experts architecture to combine multi-scale crowd density maps in a deep neural network. When grouping different experts, we allow their overlaps to drive the collaboration and competition among them. The pixel-wise soft gating net is introduced to generate pixel-wise soft weight maps for expert combination in each group. In particular, to distinguish between the focus of gating nets over two levels of the hierarchy, an attention module is inserted into the one in the second level. A novel relative local counting loss is introduced which optimizes the network from a different perspective of the density estimation. It measures the relative local count differences among hard-predicted local regions. 
%Relative Local Counting Loss enforces relatedness among different regions by calculating relative local count. 
The results on five benchmarks demonstrate that our method achieves superior performance compared with the state of the art.

\section*{Acknowledgments}
This work was supported by King's Cambridge 1 Access Fund, EPSRC IAA grant and the Fundamental Research Funds for the Central Universities.

% {\appendix[Proof of the Zonklar Equations]
% Use $\backslash${\tt{appendix}} if you have a single appendix:
% Do not use $\backslash${\tt{section}} anymore after $\backslash${\tt{appendix}}, only $\backslash${\tt{section*}}.
% If you have multiple appendixes use $\backslash${\tt{appendices}} then use $\backslash${\tt{section}} to start each appendix.
% You must declare a $\backslash${\tt{section}} before using any $\backslash${\tt{subsection}} or using $\backslash${\tt{label}} ($\backslash${\tt{appendices}} by itself
%  starts a section numbered zero.)}

%{\appendices
%\section*{Proof of the First Zonklar Equation}
%Appendix one text goes here.
% You can choose not to have a title for an appendix if you want by leaving the argument blank
%\section*{Proof of the Second Zonklar Equation}
%Appendix two text goes here.}

 \bibliographystyle{IEEEtran}
 \bibliography{egbib}

% Generated by IEEEtran.bst, version: 1.14 (2015/08/26)
\begin{thebibliography}{10}
\providecommand{\url}[1]{#1}
\csname url@samestyle\endcsname
\providecommand{\newblock}{\relax}
\providecommand{\bibinfo}[2]{#2}
\providecommand{\BIBentrySTDinterwordspacing}{\spaceskip=0pt\relax}
\providecommand{\BIBentryALTinterwordstretchfactor}{4}
\providecommand{\BIBentryALTinterwordspacing}{\spaceskip=\fontdimen2\font plus
\BIBentryALTinterwordstretchfactor\fontdimen3\font minus
  \fontdimen4\font\relax}
\providecommand{\BIBforeignlanguage}[2]{{%
\expandafter\ifx\csname l@#1\endcsname\relax
\typeout{** WARNING: IEEEtran.bst: No hyphenation pattern has been}%
\typeout{** loaded for the language `#1'. Using the pattern for}%
\typeout{** the default language instead.}%
\else
\language=\csname l@#1\endcsname
\fi
#2}}
\providecommand{\BIBdecl}{\relax}
\BIBdecl

\bibitem{zhang2016cvpr}
Y.~Zhang, D.~Zhou, S.~Chen, S.~Gao, and Y.~Ma, ``Single-image crowd counting
  via multi-column convolutional neural network,'' in \emph{CVPR}, 2016.

\bibitem{onoro2016eccv}
D.~Onoro-Rubio and R.~J. L{\'o}pez-Sastre, ``Towards perspective-free object
  counting with deep learning,'' in \emph{ECCV}, 2016.

\bibitem{sam2017cvpr}
D.~Babu~Sam, S.~Surya, and R.~Venkatesh~Babu, ``Switching convolutional neural
  network for crowd counting,'' in \emph{CVPR}, 2017.

\bibitem{ranjan2018eccv}
V.~Ranjan, H.~Le, and M.~Hoai, ``Iterative crowd counting,'' in \emph{ECCV},
  2018.

\bibitem{boominathan2016mm}
L.~Boominathan, S.~S. Kruthiventi, and R.~V. Babu, ``Crowdnet: a deep
  convolutional network for dense crowd counting,'' in \emph{ACM MM}, 2016.

\bibitem{zhang2018wacv}
Z.~Lu, M.~Shi, and Q.~Chen, ``Crowd counting via scale-adaptive convolutional
  neural network,'' in \emph{WACV}, 2018.

\bibitem{liu2019iccv}
L.~Liu, Z.~Qiu, G.~Li, S.~Liu, W.~Ouyang, and L.~Lin, ``Crowd counting with
  deep structured scale integration network,'' in \emph{ICCV}, 2019.

\bibitem{shi2019cvpr}
M.~Shi, Z.~Yang, C.~Xu, and Q.~Chen, ``Revisiting perspective information for
  efficient crowd counting,'' in \emph{CVPR}, 2019.

\bibitem{yan2019iccv}
Z.~Yan, Y.~Yuan, W.~Zuo, X.~Tan, Y.~Wang, S.~Wen, and E.~Ding,
  ``Perspective-guided convolution networks for crowd counting,'' in
  \emph{ICCV}, 2019.

\bibitem{yang2020cvpr}
Y.~Yang, G.~Li, Z.~Wu, L.~Su, Q.~Huang, and N.~Sebe, ``Reverse perspective
  network for perspective-aware object counting,'' in \emph{CVPR}, 2020.

\bibitem{shi2019iccv}
Z.~Shi, P.~Mettes, and C.~G. Snoek, ``Counting with focus for free,'' in
  \emph{ICCV}, 2019.

\bibitem{jiang2020cvpr}
X.~Jiang, L.~Zhang, M.~Xu, T.~Zhang, P.~Lv, B.~Zhou, X.~Yang, and Y.~Pang,
  ``Attention scaling for crowd counting,'' in \emph{CVPR}, 2020.

\bibitem{jacobs1991nn}
R.~A. Jacobs, M.~I. Jordan, S.~J. Nowlan, and G.~E. Hinton, ``Adaptive mixtures
  of local experts,'' \emph{Neural computation}, vol.~3, no.~1, pp. 79--87,
  1991.

\bibitem{jordan1994nn}
M.~I. Jordan and R.~A. Jacobs, ``Hierarchical mixtures of experts and the em
  algorithm,'' \emph{Neural computation}, vol.~6, no.~2, pp. 181--214, 1994.

\bibitem{collobert2002nn}
R.~Collobert, S.~Bengio, and Y.~Bengio, ``A parallel mixture of svms for very
  large scale problems,'' \emph{Neural computation}, vol.~14, no.~5, pp.
  1105--1114, 2002.

\bibitem{tresp2001nips}
V.~Tresp, ``Mixtures of gaussian processes,'' in \emph{NeurIPS}, 2001.

\bibitem{eigen2013arxiv}
D.~Eigen, M.~Ranzato, and I.~Sutskever, ``Learning factored representations in
  a deep mixture of experts,'' \emph{arXiv preprint arXiv:1312.4314}, 2013.

\bibitem{kumagai2018MVA}
S.~Kumagai, K.~Hotta, and T.~Kurita, ``Mixture of counting cnns,''
  \emph{Machine Vision and Applications}, vol.~29, no.~7, pp. 1119--1126, 2018.

\bibitem{idrees2013cvpr}
H.~Idrees, I.~Saleemi, C.~Seibert, and M.~Shah, ``Multi-source multi-scale
  counting in extremely dense crowd images,'' in \emph{CVPR}, 2013.

\bibitem{chen2012bmvc}
K.~Chen, C.~C. Loy, S.~Gong, and T.~Xiang, ``Feature mining for localised crowd
  counting.'' in \emph{Bmvc}, 2012.

\bibitem{liu2020eccv}
X.~Liu, J.~Yang, W.~Ding, T.~Wang, Z.~Wang, and J.~Xiong, ``Adaptive mixture
  regression network with local counting map for crowd counting,'' in
  \emph{ECCV}, 2020.

\bibitem{sindagi2020jhu}
V.~Sindagi, R.~Yasarla, and V.~M. Patel, ``Jhu-crowd++: Large-scale crowd
  counting dataset and a benchmark method,'' \emph{IEEE Transactions on Pattern
  Analysis and Machine Intelligence}, 2020.

\bibitem{wang2020pami}
Q.~Wang, J.~Gao, W.~Lin, and X.~Li, ``Nwpu-crowd: A large-scale benchmark for
  crowd counting and localization,'' \emph{IEEE transactions on pattern
  analysis and machine intelligence}, vol.~43, no.~6, pp. 2141--2149, 2020.

\bibitem{guerrero2015ibpra}
R.~Guerrero-G{\'o}mez-Olmedo, B.~Torre-Jim{\'e}nez, R.~L{\'o}pez-Sastre,
  S.~Maldonado-Basc{\'o}n, and D.~Onoro-Rubio, ``Extremely overlapping vehicle
  counting,'' in \emph{Iberian Conference on Pattern Recognition and Image
  Analysis}, 2015.

\bibitem{chan2009iccv}
A.~B. Chan and N.~Vasconcelos, ``Bayesian poisson regression for crowd
  counting,'' in \emph{ICCV}, 2009.

\bibitem{paul2017iccvw}
J.~Paul~Cohen, G.~Boucher, C.~A. Glastonbury, H.~Z. Lo, and Y.~Bengio,
  ``Count-ception: Counting by fully convolutional redundant counting,'' in
  \emph{ICCVw}, 2017.

\bibitem{lu2017plantmethod}
H.~Lu, Z.~Cao, Y.~Xiao, B.~Zhuang, and C.~Shen, ``Tasselnet: counting maize
  tassels in the wild via local counts regression network,'' \emph{Plant
  methods}, vol.~13, no.~1, pp. 1--17, 2017.

\bibitem{shang2016icip}
C.~Shang, H.~Ai, and B.~Bai, ``End-to-end crowd counting via joint learning
  local and global count,'' in \emph{ICIP}, 2016.

\bibitem{lempitsky2010nips}
V.~Lempitsky and A.~Zisserman, ``Learning to count objects in images,'' in
  \emph{NIPS}, 2010.

\bibitem{sindagi2017iccv}
V.~A. Sindagi and V.~M. Patel, ``Generating high-quality crowd density maps
  using contextual pyramid cnns,'' in \emph{ICCV}, 2017.

\bibitem{li2018cvpr}
Y.~Li, X.~Zhang, and D.~Chen, ``Csrnet: Dilated convolutional neural networks
  for understanding the highly congested scenes,'' in \emph{CVPR}, 2018.

\bibitem{xu2019iccv}
C.~Xu, K.~Qiu, J.~Fu, S.~Bai, Y.~Xu, and X.~Bai, ``Learn to scale: Generating
  multipolar normalized density maps for crowd counting,'' in \emph{ICCV},
  2019.

\bibitem{zhao2022tmm}
H.~Zhao, Q.~Wang, G.~Zhan, W.~Min, Y.~Zou, and S.~Cui, ``Need only one more
  point (noomp): Perspective adaptation crowd counting in complex scenes,''
  \emph{IEEE Transactions on Multimedia}, 2022.

\bibitem{du2023aaai}
Z.~Du, J.~Deng, and M.~Shi, ``Domain-general crowd counting in unseen
  scenarios,'' in \emph{AAAI}, 2023.

\bibitem{xiong2019iccv}
H.~Xiong, H.~Lu, C.~Liu, L.~Liu, Z.~Cao, and C.~Shen, ``From open set to closed
  set: Counting objects by spatial divide-and-conquer,'' in \emph{ICCV}, 2019.

\bibitem{liu2018cvpr}
J.~Liu, C.~Gao, D.~Meng, and A.~G.~Hauptmann, ``Decidenet: Counting varying
  density crowds through attention guided detection and density estimation,''
  in \emph{CVPR}, 2018.

\bibitem{lian2019cvpr}
D.~Lian, J.~Li, J.~Zheng, W.~Luo, and S.~Gao, ``Density map regression guided
  detection network for rgb-d crowd counting and localization,'' in
  \emph{CVPR}, 2019.

\bibitem{liushi2019cvpr}
Y.~Liu, M.~Shi, Q.~Zhao, and X.~Wang, ``Point in, box out: Beyond counting
  persons in crowds,'' in \emph{CVPR}, 2019.

\bibitem{liu2020acmmm}
Y.~Liu, Z.~Wang, M.~Shi, S.~Satoh, Q.~Zhao, and H.~Yang, ``Towards unsupervised
  crowd counting via regression-detection bi-knowledge transfer,'' in \emph{ACM
  MM}, 2020.

\bibitem{zhang2021cvpr}
Q.~Zhang, W.~Lin, and A.~B. Chan, ``Cross-view cross-scene multi-view crowd
  counting,'' in \emph{CVPR}, 2021.

\bibitem{qiu2019icme}
Z.~Qiu, L.~Liu, G.~Li, Q.~Wang, N.~Xiao, and L.~Lin, ``Crowd counting via
  multi-view scale aggregation networks,'' in \emph{ICME}, 2019.

\bibitem{liu2021cvpr}
L.~Liu, J.~Chen, H.~Wu, G.~Li, C.~Li, and L.~Lin, ``Cross-modal collaborative
  representation learning and a large-scale rgbt benchmark for crowd
  counting,'' in \emph{CVPR}, 2021.

\bibitem{liu2018bcvpr}
X.~Liu, J.~Weijer, and A.~D. Bagdanov, ``Leveraging unlabeled data for crowd
  counting by learning to rank,'' in \emph{CVPR}, 2018.

\bibitem{cheng2019iccv}
Z.-Q. Cheng, J.-X. Li, Q.~Dai, X.~Wu, and A.~G. Hauptmann, ``Learning spatial
  awareness to improve crowd counting,'' in \emph{ICCV}, 2019.

\bibitem{zhao2020eccv}
Z.~Zhao, M.~Shi, X.~Zhao, and L.~Li, ``Active crowd counting with limited
  supervision,'' in \emph{ECCV}, 2020.

\bibitem{yang2020eccv}
Y.~Yang, G.~Li, Z.~Wu, L.~Su, Q.~Huang, and N.~Sebe, ``Weakly-supervised crowd
  counting learns from sorting rather than locations,'' in \emph{ECCV}, 2020.

\bibitem{cao2018eccv}
X.~Cao, Z.~Wang, Y.~Zhao, and F.~Su, ``Scale aggregation network for accurate
  and efficient crowd counting,'' in \emph{ECCV}, 2018.

\bibitem{mo2022tip}
H.~Mo, W.~Ren, X.~Zhang, F.~Yan, Z.~Zhou, X.~Cao, and W.~Wu, ``Attention-guided
  collaborative counting,'' \emph{IEEE Transactions on Image Processing},
  vol.~31, pp. 6306--6319, 2022.

\bibitem{shazeer2017arxiv}
N.~Shazeer, A.~Mirhoseini, K.~Maziarz, A.~Davis, Q.~Le, G.~Hinton, and J.~Dean,
  ``Outrageously large neural networks: The sparsely-gated mixture-of-experts
  layer,'' \emph{arXiv preprint arXiv:1701.06538}, 2017.

\bibitem{zhang2019cvpr}
L.~Zhang, S.~Huang, W.~Liu, and D.~Tao, ``Learning a mixture of
  granularity-specific experts for fine-grained categorization,'' in
  \emph{CVPR}, 2019.

\bibitem{shi2019arxiv}
Y.~Shi, N.~Siddharth, B.~Paige, and P.~H. Torr, ``Variational
  mixture-of-experts autoencoders for multi-modal deep generative models,''
  \emph{arXiv preprint arXiv:1911.03393}, 2019.

\bibitem{ma2018kdd}
J.~Ma, Z.~Zhao, X.~Yi, J.~Chen, L.~Hong, and E.~H. Chi, ``Modeling task
  relationships in multi-task learning with multi-gate mixture-of-experts,'' in
  \emph{SIGKDD}, 2018.

\bibitem{liu2021iccv}
X.~Liu, G.~Li, Z.~Han, W.~Zhang, Y.~Yang, Q.~Huang, and N.~Sebe, ``Exploiting
  sample correlation for crowd counting with multi-expert network,'' in
  \emph{ICCV}, 2021.

\bibitem{sam2018cvpr}
D.~B. Sam, N.~N. Sajjan, R.~V. Babu, and M.~Srinivasan, ``Divide and grow:
  Capturing huge diversity in crowd images with incrementally growing cnn,'' in
  \emph{CVPR}, 2018.

\bibitem{shi2018cvpr}
Z.~Shi, L.~Zhang, Y.~Liu, X.~Cao, Y.~Ye, M.-M. Cheng, and G.~Zheng, ``Crowd
  counting with deep negative correlation learning,'' in \emph{CVPR}, 2018.

\bibitem{zhang2021pami}
L.~Zhang, Z.~Shi, M.-M. Cheng, Y.~Liu, J.-W. Bian, J.~T. Zhou, G.~Zheng, and
  Z.~Zeng, ``Nonlinear regression via deep negative correlation learning,''
  \emph{IEEE Transactions on Pattern Analysis and Machine Intelligence},
  vol.~43, no.~3, pp. 982--998, 2021.

\bibitem{ng2014arxiv}
J.~W. Ng and M.~P. Deisenroth, ``Hierarchical mixture-of-experts model for
  large-scale gaussian process regression,'' \emph{arXiv preprint
  arXiv:1412.3078}, 2014.

\bibitem{chen2022nips}
Z.~Chen, Y.~Deng, Y.~Wu, Q.~Gu, and Y.~Li, ``Towards understanding the
  mixture-of-experts layer in deep learning,'' \emph{NeurIPS}, 2022.

\bibitem{fedus2021jmlr}
W.~Fedus, B.~Zoph, and N.~Shazeer, ``Switch transformers: Scaling to trillion
  parameter models with simple and efficient sparsity,'' \emph{J. Mach. Learn.
  Res}, vol.~23, pp. 1--40, 2021.

\bibitem{riquelme2021nips}
C.~Riquelme, J.~Puigcerver, B.~Mustafa, M.~Neumann, R.~Jenatton,
  A.~Susano~Pinto, D.~Keysers, and N.~Houlsby, ``Scaling vision with sparse
  mixture of experts,'' \emph{NeurIPS}, 2021.

\bibitem{wang2021iccv}
C.~Wang, Q.~Song, B.~Zhang, Y.~Wang, Y.~Tai, X.~Hu, C.~Wang, J.~Li, J.~Ma, and
  Y.~Wu, ``Uniformity in heterogeneity: Diving deep into count interval
  partition for crowd counting,'' in \emph{ICCV}, 2021.

\bibitem{rong2021wacv}
L.~Rong and C.~Li, ``Coarse-and fine-grained attention network with
  background-aware loss for crowd density map estimation,'' in \emph{WACV},
  2021.

\bibitem{guerrero2015icpria}
R.~Guerrero-G{\'o}mez-Olmedo, B.~Torre-Jim{\'e}nez, R.~L{\'o}pez-Sastre,
  S.~Maldonado-Basc{\'o}n, and D.~Onoro-Rubio, ``Extremely overlapping vehicle
  counting,'' in \emph{ICPRIA}, 2015.

\bibitem{tian2019tip}
Y.~Tian, Y.~Lei, J.~Zhang, and J.~Z. Wang, ``Padnet: Pan-density crowd
  counting,'' \emph{IEEE Transactions on Image Processing}, vol.~29, pp.
  2714--2727, 2019.

\bibitem{wang2020nips}
B.~Wang, H.~Liu, D.~Samaras, and M.~Hoai, ``Distribution matching for crowd
  counting,'' in \emph{NeurIPS}, 2020.

\bibitem{sam2020pami}
D.~B. Sam, S.~V. Peri, M.~N. Sundararaman, A.~Kamath, and V.~B. Radhakrishnan,
  ``Locate, size and count: Accurately resolving people in dense crowds via
  detection,'' \emph{IEEE transactions on pattern analysis and machine
  intelligence}, 2020.

\bibitem{liu2020tip}
Y.~Liu, Q.~Wen, H.~Chen, W.~Liu, J.~Qin, G.~Han, and S.~He, ``Crowd counting
  via cross-stage refinement networks,'' \emph{IEEE Transactions on Image
  Processing}, vol.~29, pp. 6800--6812, 2020.

\bibitem{wan2021cvpr}
J.~Wan, Z.~Liu, and A.~B. Chan, ``A generalized loss function for crowd
  counting and localization,'' in \emph{CVPR}, 2021.

\bibitem{chen2021iccv}
B.~Chen, Z.~Yan, K.~Li, P.~Li, B.~Wang, W.~Zuo, and L.~Zhang, ``Variational
  attention: Propagating domain-specific knowledge for multi-domain learning in
  crowd counting,'' in \emph{ICCV}, 2021.

\bibitem{cheng2021tip}
J.~Cheng, H.~Xiong, Z.~Cao, and H.~Lu, ``Decoupled two-stage crowd counting and
  beyond,'' \emph{IEEE Transactions on Image Processing}, vol.~30, pp.
  2862--2875, 2021.

\bibitem{shu2022cvpr}
W.~Shu, J.~Wan, K.~C. Tan, S.~Kwong, and A.~B. Chan, ``Crowd counting in the
  frequency domain,'' in \emph{CVPR}, 2022.

\bibitem{cheng2022cvpr}
Z.-Q. Cheng, Q.~Dai, H.~Li, J.~Song, X.~Wu, and A.~G. Hauptmann, ``Rethinking
  spatial invariance of convolutional networks for object counting,'' in
  \emph{CVPR}, 2022.

\bibitem{xiong2022eccv}
H.~Xiong and A.~Yao, ``Discrete-constrained regression for local counting
  models,'' \emph{arXiv preprint arXiv:2207.09865}, 2022.

\bibitem{simonyan2015iclr}
K.~Simonyan and A.~Zisserman, ``Very deep convolutional networks for
  large-scale image recognition,'' in \emph{ICLR}, 2015.

\bibitem{jiang2020tmm}
X.~Jiang, L.~Zhang, T.~Zhang, P.~Lv, B.~Zhou, Y.~Pang, M.~Xu, and C.~Xu,
  ``Density-aware multi-task learning for crowd counting,'' \emph{IEEE
  Transactions on Multimedia}, vol.~23, pp. 443--453, 2020.

\bibitem{yan2021tmm}
Z.~Yan, R.~Zhang, H.~Zhang, Q.~Zhang, and W.~Zuo, ``Crowd counting via
  perspective-guided fractional-dilation convolution,'' \emph{IEEE Transactions
  on Multimedia}, 2021.

\bibitem{he2016cvpr}
K.~He, X.~Zhang, S.~Ren, and J.~Sun, ``Deep residual learning for image
  recognition,'' in \emph{CVPR}, 2016.

\bibitem{zhao2015cvpr}
F.~Zhao, Y.~Huang, L.~Wang, and T.~Tan, ``Deep semantic ranking based hashing
  for multi-label image retrieval,'' in \emph{CVPR}, 2015.

\bibitem{gordo2016eccv}
A.~Gordo, J.~Almaz{\'a}n, J.~Revaud, and D.~Larlus, ``Deep image retrieval:
  Learning global representations for image search,'' in \emph{ECCV}, 2016.

\bibitem{liu2019tpami}
X.~Liu, J.~Van De~Weijer, and A.~D. Bagdanov, ``Exploiting unlabeled data in
  cnns by self-supervised learning to rank,'' \emph{IEEE transactions on
  pattern analysis and machine intelligence}, 2019.

\end{thebibliography}
 % argument is your BibTeX string definitions and bibliography database(s)
%\bibliography{IEEEabrv,../bib/paper}
%

% \newpage

% \section{Biography Section}
% If you have an EPS/PDF photo (graphicx package needed), extra braces are
%  needed around the contents of the optional argument to biography to prevent
%  the LaTeX parser from getting confused when it sees the complicated
%  $\backslash${\tt{includegraphics}} command within an optional argument. (You can create
%  your own custom macro containing the $\backslash${\tt{includegraphics}} command to make things
%  simpler here.)
 
% \vspace{11pt}

% \bf{If you include a photo:}\vspace{-33pt}
% \begin{IEEEbiography}[{\includegraphics[width=1in,height=1.25in,clip,keepaspectratio]{fig1}}]{Michael Shell}
% Use $\backslash${\tt{begin\{IEEEbiography\}}} and then for the 1st argument use $\backslash${\tt{includegraphics}} to declare and link the author photo.
% Use the author name as the 3rd argument followed by the biography text.
% \end{IEEEbiography}

% \vspace{11pt}

% \bf{If you will not include a photo:}\vspace{-33pt}
% \begin{IEEEbiographynophoto}{John Doe}
% Use $\backslash${\tt{begin\{IEEEbiographynophoto\}}} and the author name as the argument followed by the biography text.
% \end{IEEEbiographynophoto}

\vfill

\end{document}